\pdfoutput=1
\documentclass[runningheads]{llncs}

\PassOptionsToPackage{dvipsnames,svgnames,x11names,table}{xcolor}

\usepackage{eccv}

\usepackage{graphicx}
\usepackage{booktabs}

\usepackage{glossaries}
\usepackage{graphicx}
\graphicspath{{figures/}}
\usepackage{float}
\usepackage{tabularx}
\usepackage{amsmath,amssymb,amsfonts,nicefrac}
\usepackage[linesnumbered,ruled,vlined]{algorithm2e}

\SetAlFnt{\small}
\SetAlCapFnt{\small}
\SetAlCapNameFnt{\small}
\SetAlCapHSkip{0pt}

\usepackage[breaklinks,hidelinks]{hyperref}
\usepackage[capitalise,capitalize,noabbrev,nameinlink]{cleveref}
\usepackage{acronym}

\usepackage{enumitem}
\usepackage{balance}
\usepackage{xspace}
\usepackage{setspace}
\usepackage{wrapfig}
\usepackage{placeins}

\usepackage{caption}
\usepackage{subcaption}
\usepackage{booktabs,colortbl,multirow,array,makecell,tabularray}
\usepackage{overpic}
\usepackage[misc]{ifsym}
\usepackage{algorithmicx}
\usepackage{algpseudocode}

\usepackage{orcidlink}

\usepackage{silence}
\makeatletter
\DeclareRobustCommand\onedot{\futurelet\@let@token\@onedot}
\def\@onedot{\ifx\@let@token.\else.\null\fi\xspace}
\def\eg{\emph{e.g}\onedot}

\def\ie{\emph{i.e}\onedot}

\makeatother

\acrodef{vlm}[VLM]{Vision-Language Model}
\acrodef{cvae}[cVAE]{conditional Variational Auto-Encoder}
\acrodef{icp}[ICP]{Iterative Closest Points}
\acrodef{mocap}[MoCap]{motion-captured}
\acrodef{vit}[ViT]{Vision Transformer}
\acrodef{fid}[FID]{Fréchet Inception Distance}
\acrodef{gs}[GS]{Gaussian Splatting}
\acrodef{3dgs}[3DGS]{3D Gaussian Splatting}
\acrodef{nerf}[NeRF]{Neural Radiance Fields}
\acrodef{mpm}[MPM]{Material Point Method}
\acrodef{fem}[FEM]{Finite Element Method}
\acrodef{sph}[SPH]{Smoothed Particle Hydrodynamics}
\acrodef{flip}[FLIP]{Fluid-Implicit-Particle}
\acrodef{pic}[PIC]{Particle-in-Cell}
\acrodef{fp}[FP]{FLIP/PIC Blend}
\acrodef{xfem}[XFEM]{Extended Finite Element Method}
\acrodef{bem}[BEM]{Boundary Element Method}
\acrodef{sds}[SDS]{Score Distillation Sampling}
\acrodef{sh}[SH]{Spherical Harmonic}
\acrodef{dssim}[D-SSIM]{Structural Dissimilarity Index Measure}
\acrodef{cdm}[CDM]{Continuum Damage Mechanics}
\acrodef{cdmpm}[CD-MPM]{Continuum Damage Material Point Method}
\acrodef{pff}[PFF]{Phase-Field Fracture}
\acrodef{nacc}[NACC]{Non-Associated Cam-Clay}
\acrodef{fm}[FM]{Fracture Mechanics}
\acrodef{pca}[PCA]{Principal Component Analysis}
\acrodef{pbr}[PBR]{Physically Based Rendering}

\newcommand{\modelname}{\textbf{AutoPhyX}\xspace}
\newcommand{\dataset}{\textbf{Text2PhyX}\xspace}
\newcommand{\supp}{\textit{Supplementary Material}\xspace}

\makeatletter
\renewcommand{\paragraph}{%
  \@startsection{paragraph}{4}{\z@}%
  {1ex plus 0.5ex minus 0.2ex} %
  {-1em}                      %
  {\normalfont\normalsize\bfseries} %
}
\makeatother

\begin{document}

\title{AutoPhyX: Automatic Text-Condition Physics Property Generation}

\titlerunning{AutoPhyX: Automatic Text-Condition Physics Property Generation}

\author{Bei Huang\inst{1,2}\thanks{Work done as an intern at BIGAI.~~~$^{\dagger}$Project Lead.~~~\Letter~Corresponding authors.}\orcidlink{0009-0006-3634-4157},
Yixin Chen\inst{2}$^{\dagger\text{\Letter}}$\orcidlink{0000-0002-8176-0241},
Hongbin Zha\inst{1,3},\\
Yuru Pei\inst{1}$^\text{\Letter}$,
Siyuan Huang\inst{2}$^\text{\Letter}$\orcidlink{0000-0003-1524-7148}}

\authorrunning{B.~Huang et al.}

\institute{State Key Laboratory of GAI, School of IST, Peking University, China \and
State Key Laboratory of General Artificial Intelligence, BIGAI, China \and
School of Artificial Intelligence and Computer Science, Anqing Normal University\\[2mm]
\textbf{\href{https://hb-pencil-zero.github.io/AutoPhyX}{hb-pencil-zero.github.io/AutoPhyX}}}

\maketitle

\begin{abstract}
High-fidelity 4D physical simulation requires accurate physical parameter assignments that align with diverse material properties. Previous methods struggle with critical bottlenecks: they either rely on computationally intensive per-scene optimization or predict parameters from purely visual inputs. The latter suffers from inherent ambiguity, \eg, the inability to distinguish the stiffness of rubber from its appearance alone.
To bridge this gap, we propose \modelname, a text-conditioned framework for predicting spatially-varying physical properties. Since training robust feed-forward models requires fine-grained supervision, we first introduce a novel part-controllable data generation pipeline that decomposes complex 3D assets into semantically distinct components and pairs them with physically plausible, diverse parameters and corresponding text descriptions.
Leveraging this dataset, \modelname employs a cross-modal modulation mechanism where text features dynamically modulate visual features for precise physical grounding. By formulating property prediction within a voxel field, \modelname ensures compatibility with diverse 3D representations, including meshes, point clouds, \acl{gs}, and \acs{nerf}. Experiments demonstrate that our method enables physically plausible, text-driven parameter assignment in a single forward pass. It achieves high accuracy and robust generalization to in-the-wild objects, paving the way for future physics-intensive embodied and robotic applications.
\end{abstract}

\section{Introduction}
\label{sec:intro}
High-fidelity 4D physical simulation ~\cite{wu20244d,mou2024instruct,shao2023tensor4d,luo2025vr} has become a vital foundation for interactive digital environments, driving applications ranging from \ac{fem} based~\cite{moes1999finite, bathe2007finite, baraff1997introduction,bathe1975finite, belytschko2014nonlinear} robotic policy training, to \ac{sph} driven~\cite{monaghan1992smoothed, monaghan2012smoothed, monaghan2005smoothed, morris1996analysis,swegle1995smoothed,randles1996smoothed,liu2006restoring} complex fluid and glacier calving simulations, and \ac{mpm} powered~\cite{de2020material,bardenhagen2004generalized,jiang2016material,wolper2019cd, solowski2015evaluation, chen2002phase, chen2026empm, wolper2019cd, wolper2020anisompm, wolper2021glacier} real-time virtual reality (VR) experiences. Despite their ability to synthesize high-fidelity, realistic dynamics across diverse scenes, these physics-based approaches share a major bottleneck: they rely heavily on the manual assignment of physical parameters. This reliance on trial-and-error parameter design severely limits their scalability and applicability to complex scenes characterized by diverse, spatially varying materials.

\begin{figure}[t!]
  \centering
  \includegraphics[width=\textwidth]{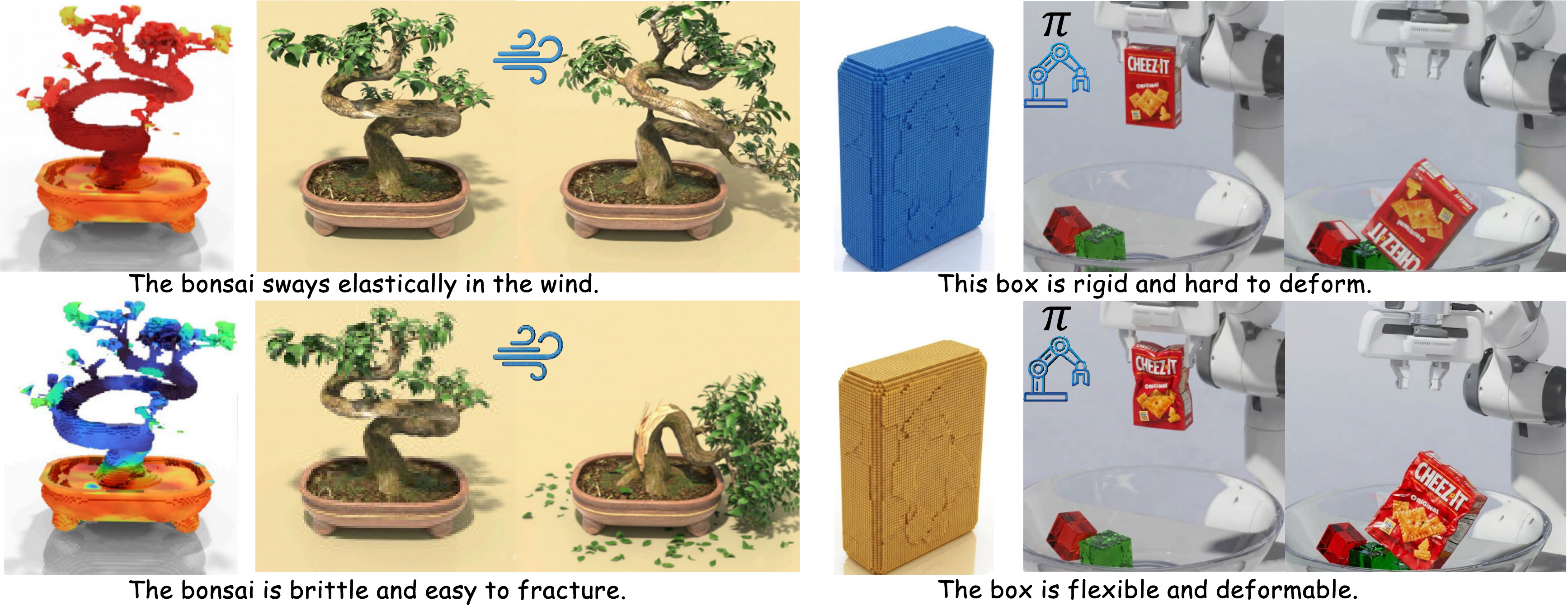}
  \caption{\textbf{Text-driven physical simulation with \modelname.} Our framework predicts dense physical parameters for 3D assets conditioned on textual descriptions. For each scenario, the leftmost images visualize the predicted voxelized physical parameters, which are normalized and mapped to RGB channels, corresponding to the input text. Under identical boundary conditions, the resulting simulations exhibit distinct behaviors, \eg, the bonsai responding differently to the same wind force in MPM, and the box with different deformation under the same robot grasping policy in FEM.}
  \label{fig:teaser}
\end{figure}

Existing methods aiming to automate material acquisition primarily fall into two categories, each with its own limitations. Optimization-based methods~\cite{lin2025omniphysgs,huang2025dreamphysics,zhang2024physdreamer,zhai2024physical} leverage differentiable physics to iteratively refine material fields to match target observations. However, inferring particle-wise parameters from sparse supervision is ill-posed and computationally expensive, often requiring hours of per-scene optimization~\cite{lin2025omniphysgs} and offering limited generalization across scenes. In contrast, feed-forward approaches such as Pixie~\cite{le2025pixiefastgeneralizablesupervised} and VoMP~\cite{dagli2025vomppredictingvolumetricmechanical} enable much faster inference; yet they rely purely on visual cues, which sacrifices semantic controllability and suffers from appearance ambiguity. For example, as shown in ~\cref{fig:teaser}, a tree could feature flexible branches that sway in the wind or brittle wood that fractures easily, while a biscuit box could be rigid cardboard or thin paper. Without explicit semantic guidance, it is difficult to disambiguate objects that are physically distinct but visually similar.

To address these critical limitations, we propose \modelname, a feed-forward framework for fast, generalizable, and semantically controllable prediction of physical properties. Given a 3D asset and textual descriptions of its physical properties, the model predicts spatially dense physical parameters in a unified voxel field representation. This formulation provides a common physical abstraction that can be directly transferred to diverse downstream 3D formats, including \ac{gs}, meshes, and point clouds, without per-scene optimization. To resolve the material-level ambiguity and capture spatially varying properties across multi-part, multi-material assets, \modelname modulates the intermediate visual representation with a feature-wise affine transformation conditioned on the encoded text features, coupling high-level semantic cues with fine-grained spatial representations. This cross-modal design enables the model to distinguish visually similar but physically different materials, while also assigning spatially coherent physical parameters to semantically distinct parts of the 3D asset.

Training such a robust, spatially-grounded model requires large-scale fine-grained supervision over multi-part, multi-material 3D assets with text descriptions, which is currently unavailable. To bridge this gap, we introduce an automated, part-controllable data generation pipeline. For each asset, we construct a 3D OpenCLIP~\cite{ilharco_gabriel_2021_5143773} voxel representation, and leverage authoritative engineering databases (\eg, MatWeb and Wikipedia) and \acp{vlm} to extract robust text-guided part segmentation, assign plausible material properties, and generate diverse part-aware textual descriptions. Our dataset, \dataset, features 1,700 high-quality 3D assets, each paired with 8 distinct physical-parameter and text-description annotations, supplementing as training and testing data for text-conditioned physical property assignment.

\looseness=-1
We evaluate \modelname extensively on \dataset and the real-world ABO-500 ~\cite{zhai2024physical, collins2022abo} dataset, achieving state-of-the-art quantitative results on both. Specifically, our method improves the PSNR from $23.421$ to $26.012$ on our dataset and reduces the Average Displacement Error (ADE) by $20\%$ compared to the second-best method on ABO-500.
Furthermore, our predicted properties support a wide range of physical simulators, which we demonstrate through extensive results using both MPM and \ac{fem}. This versatility allows us to flexibly choose rendering and simulation pipelines on demand, \eg, \ac{gs} for real-time dynamic rendering, mesh rendering in Blender, or Isaac Gym for robotic policy training.

In summary, our main contributions are:
\begin{itemize}
    \item \textbf{The first text-conditioned feed-forward framework for 3D physics prediction.} We introduce \modelname that achieves precise semantic and physical grounding through a novel cross-modal modulation architecture, resolving the inherent ambiguity of previous pure-vision methods.
    \item \textbf{A large-scale, multi-part, multi-level 3D physics dataset.} We build \dataset with dense part-aware physical annotations and paired text descriptions, enabling supervised learning for text-conditioned property prediction with an automated \acs{vlm}-driven generation pipeline.
    \item \textbf{Extensive validation across diverse workflows.} We demonstrate that \modelname significantly outperforms existing baselines in physical accuracy. More importantly, our approach bridges the gap between static 3D assets and various physics solvers, providing a plug-and-play solution that generalizes across varied representations, platforms, and rendering engines.

\end{itemize}

\section{Related Work}
Physical simulation has long been a cornerstone of computer graphics
. Typical simulation methods, \eg, \ac{fem}, which models stress-strain responses for rigid and soft bodies, and \ac{mpm}~\cite{jiang2016material,de2020material}, particularly suited for large deformations and complex materials, have been widely adopted. Despite their success, they all rely on accurate physical parameters as input. In this work, we are particularly interested in three canonical properties: Young's modulus $E$, Poisson's ratio $\nu$, and density $\rho$, because they jointly govern a material's stiffness, lateral deformation behavior, and inertial response, making them fundamental for a broad range of solid and deformable simulations. To eliminate the need for manual parameter assignment, prior efforts can be broadly grouped into the following directions.

\subsection{Physics Property Assignment via \acl{vlm}}
Prior research, such as NeRF2Physics~\cite{zhai2024physical}, directly prompts \ac{vlm} to predict exact physical parameters from images~\cite{chen2025physgen3d,zhai2024physical,hsu2025autovfx,cao2025physx }. However, this direct approach struggles for three main reasons: \ac{vlm} lacks accurate built-in physics knowledge, their outputs are highly random, and materials suffer from severe visual ambiguity, where objects that look identical can have completely different physical properties. To overcome these issues, we shift the \ac{vlm}'s role from a direct physics calculator to a semantic reasoning engine supported by external tools. Specifically, instead of relying on the \ac{vlm} to guess exact numbers, we give specific material requirements, \eg, more deformable or crisp, and prompt it to query external databases to find reasonable physical ranges. To handle the randomness of \ac{vlm} outputs, we query the model multiple times and select the best result. Finally, we use the \ac{vlm} to generate text descriptions that explicitly define the material, effectively solving the visual ambiguity problem.

\subsection{Test-time Physics Property Optimization}
Another prominent line of research focuses on test-time optimization, which iteratively refines physical parameters to align simulations with target observations. Observation-based optimization techniques~\cite{murthy2020gradsim,li2023pac,zhong2024reconstruction,jiang2025phystwin,zhang2025particle, xiong2025topogaussian, jiang2025phystwin, ma2026fastphysgs, zhan2026perpetualwonder, li2025wonderplay} typically minimize the discrepancy between simulated deformations and ground-truth data, such as multi-view videos or particle trajectories under known forces. While effective, these methods often require high-quality, dense supervision that is difficult to acquire in real-world settings.

To alleviate this dependency, approaches~\cite{huang2025dreamphysics,zhang2024physdreamer,lin2025omniphysgs} propose leveraging pretrained video diffusion models to guide physics discovery via motion distillation losses. Notable examples include PhysDreamer~\cite{zhang2024physdreamer} and DreamPhysics~\cite{huang2025dreamphysics}, which optimize material fields by ensuring simulated dynamics remain consistent with realistic motion priors. Despite their ability to discover physics from sparse inputs, these optimization-based paradigms suffer from extensive computational overhead, often requiring hours of optimization to resolve hundreds of thousands of material points. Furthermore, these methods are characterized by heavy per-scene memorization; for each new scenario, the costly optimization must be executed from scratch, failing to generalize across diverse scenes.

\subsection{ Feedforward  Physics Property Prediction}
Feedforward methods such as Pixie~\cite{le2025pixiefastgeneralizablesupervised} and VoMP~\cite{dagli2025vomppredictingvolumetricmechanical} have recently emerged to bypass the computational burdens of test-time optimization by directly regressing physical attributes in a single pass. However, these methods rely exclusively on visual cues, which inherently lead to physical ambiguity, \ie, visually similar objects, such as fresh fruit and its plastic replica, can possess drastically different material properties. To resolve this limitation, \modelname introduces textual constraints as a crucial semantic prior to resolve the visual ambiguities, enabling the differentiation of materials with similar appearances but distinct physical behaviors.
Unlike previous models, our goal is to ensure that the predicted parameters are efficient to compute, controllable, and physically plausible, providing a versatile solution for augmenting diverse 3D representations, including meshes, point clouds, and \acl{gs}.

\section{Method}
\label{sec:method}
\begin{figure}[t!]
    \centering
    \includegraphics[width=\linewidth]{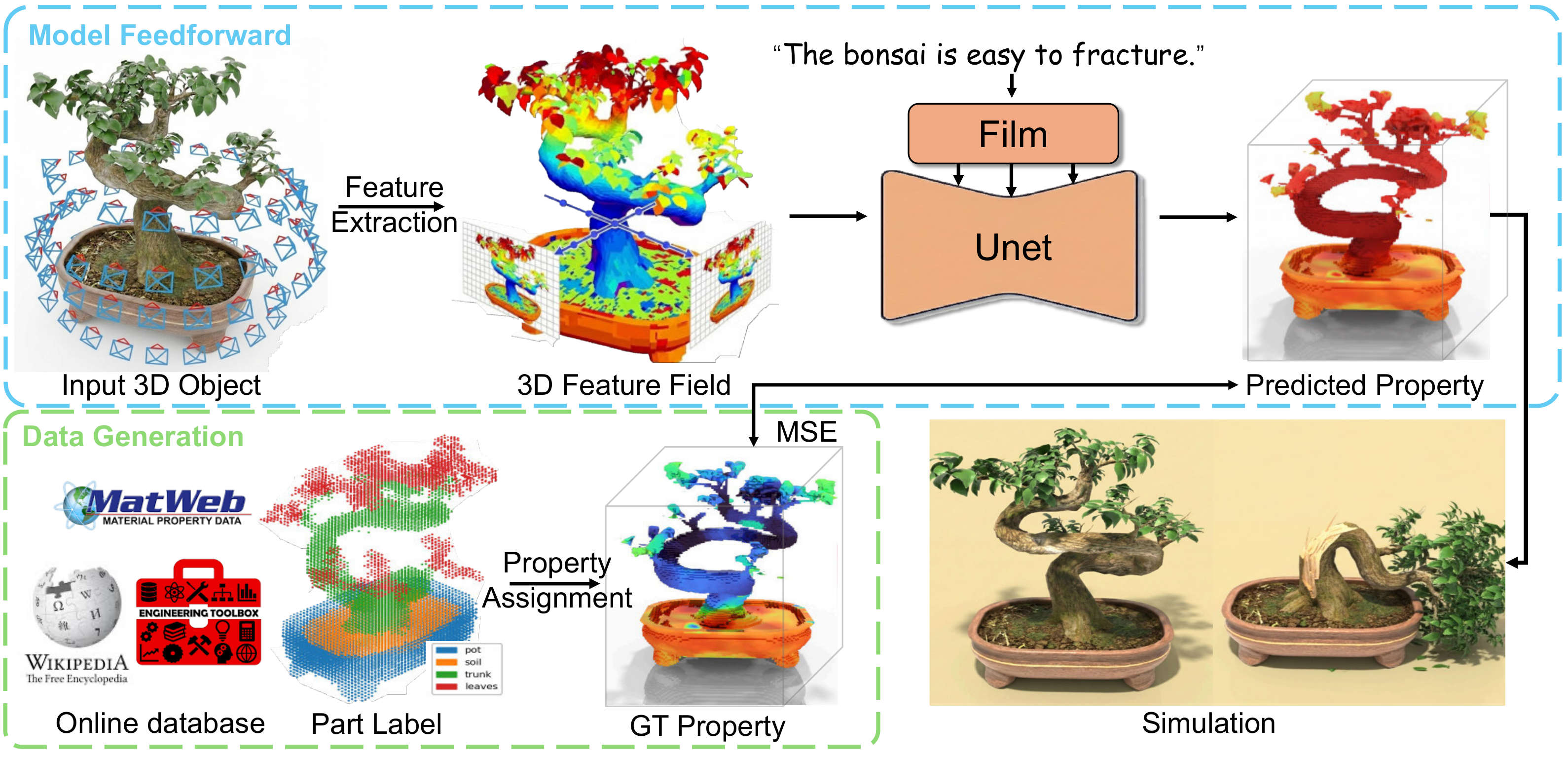}
    \caption{\textbf{Overview of our pipeline.}  Given a 3D object, our framework first decomposes it into semantically distinct components via multi-view feature rendering. Then, a text-conditioned neural network resolves visual ambiguity through cross-modal modulation, predicting volumetric physical properties in a single forward pass. The predicted properties are formulated as a voxel field, ensuring compatibility with diverse 3D representations, \eg, meshes, point clouds, and Gaussian Splatting, for physics simulation.}
    \label{fig:workflow}
\end{figure}

 Given a 3D object represented with multi-view images and a text prompt, our goal is to predict a dense, physics-ready volumetric field of material parameters, specifically Young's modulus $E$, Poisson's ratio $\nu$, and density $\rho$. To achieve this, we propose \modelname that operates in two primary stages. First, it extracts 2D OpenCLIP features from multi-view images and lifts them into a dense 3D OpenCLIP voxel grid, explicitly handling self-occlusion via volume rendering and filling the unobservable interior regions to ensure a solid physical volume (~\cref{sec:get3dfeat}). Second, we fuse these 3D visual features with textual embeddings to resolve inherent visual ambiguities, directly decoding this cross-modal representation into a unified voxel field of precise physical parameters ready for downstream simulation (\cref{sec:physics_learning}). We additionally generate a large-scale dataset to facilitate training and testing, as detailed in \cref{sec:data}.

\subsection{3D Feature Extraction from 2D Images}
\label{sec:get3dfeat}

\begin{figure}[t]
    \centering
    \subfloat[\textbf{3D Feature Extraction Comparison}\label{fig:extract3d}]{
        \begin{minipage}[c]{0.58\textwidth}
            \centering
            \includegraphics[width=\textwidth]{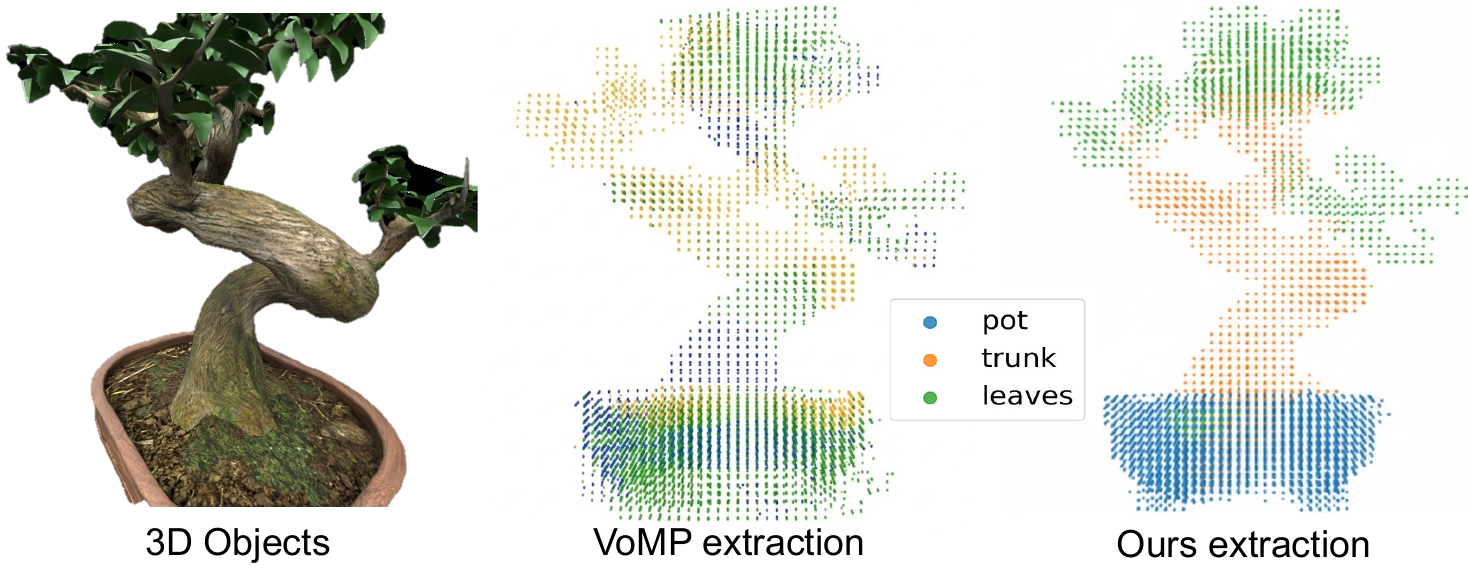}
        \end{minipage}
    }
    \hfill
    \subfloat[\textbf{Internal Voxel Filling}\label{fig:filling}]{
        \begin{minipage}[c]{0.38\textwidth}
            \centering
            \includegraphics[width=\textwidth]{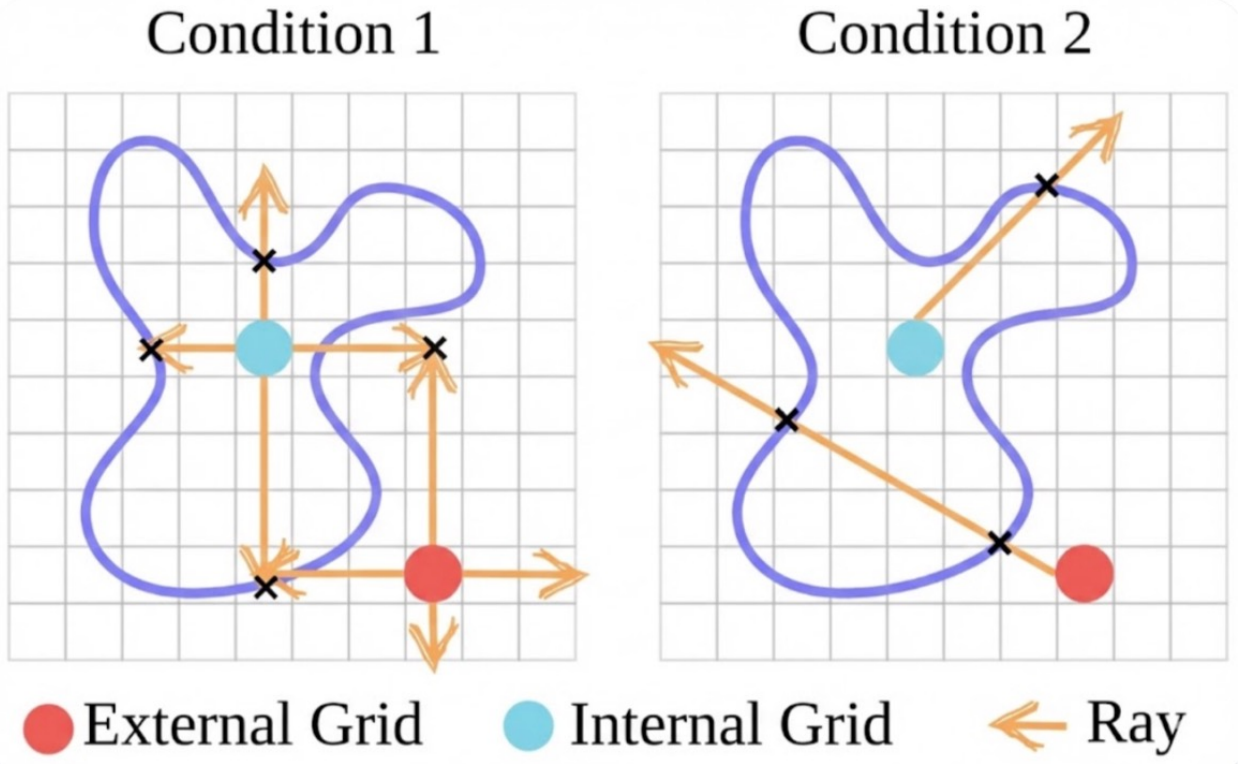}
        \end{minipage}
    }

    \caption{\textbf{Feature extraction and volumetric filling.}
    \textbf{(a)} Unlike direct averaging (\eg, VoMP~\cite{dagli2025vomppredictingvolumetricmechanical}), which suffers from feature contamination by ignoring visibility, our \ac{nerf}-based extraction preserves 3D structural integrity and spatial consistency.
    \textbf{(b)} Hierarchical classification for internal voxel filling. Condition 1 performs coarse enclosure detection via six-direction boundary checking. Condition 2 provides topological refinement using ray-surface intersection counts to filter candidates and achieve accurate filling in complex manifolds.}
    \label{fig:feature_pipeline}
\end{figure}

Recent methods like VoMP~\cite{dagli2025vomppredictingvolumetricmechanical} lift 2D features to 3D via naive multi-view averaging. However, ignoring visibility causes severe feature contamination, such as foreground leaves erroneously projecting features onto an occluded trunk, as shown in~\cref{fig:feature_pipeline}. To address these geometric ambiguities, we shift from heuristic projection to a volumetric rendering framework. We treat the 2D OpenCLIP~\cite{ilharco_gabriel_2021_5143773} features as ``feature radiance'' and optimize a continuous 3D feature field $\Phi: \mathbb{R}^3 \to \mathbb{R}^D$ via end-to-end differentiable rendering. Using a fixed density field $\sigma$ from a pre-trained RGB \ac{nerf}, we render the expected 2D feature $\hat{F}(\mathbf{r})$ along a camera ray $\mathbf{r}(t) = \mathbf{o} + t\mathbf{d}$ as
$\hat{F}(\mathbf{r}) = \int_{t_n}^{t_f} T(t) \sigma(\mathbf{r}(t)) \Phi(\mathbf{r}(t)) dt,$
where $t_n$ and $t_f$ denote the near and far bounds of the ray, and $T(t) = \exp(-\int_{t_n}^{t} \sigma(\mathbf{r}(s)) ds)$ represents the accumulated transmittance. Here, $T(t)$ naturally suppresses occluded regions by dropping to zero behind surfaces, ensuring that the 3D feature field $\Phi$ is supervised from viewpoints where the surfaces are truly visible.

While volume rendering captures surface features, physics-based simulations inherently require volumetric properties, such as $(E, \nu, \rho)$, defined throughout the object's interior to accurately model stress propagation. To bridge this gap, we fill unobservable internal voids using a hierarchical ray-casting strategy inspired by PhysGaussian~\cite{xie2024physgaussian}.
Specifically, following the classic even-odd rule in computer graphics, we identify internal voxels by casting multiple rays to infinity, where a consistently odd number of surface intersections across all rays indicates the voxel is inside, robustly handling complex concavities.
Finally, these voxels inherit semantic features via nearest-neighbor interpolation from the surface, yielding a physically and semantically consistent volumetric representation.

\subsection{Text-conditioned Physical Parameter Prediction}
\label{sec:physics_learning}

While vision-based 3D representations, such as 3D CLIP feature field~\cite{shen2023F3RM}, have proven highly effective for open-vocabulary localization and commonsense reasoning, relying solely on pure vision is fundamentally insufficient for predicting physical properties. This limitation arises from the inherent ambiguity of visual appearance. For instance, a rubber toy might look identical whether it is extremely stiff or highly deformable. Therefore, it is essential to introduce textual descriptions as a disambiguating condition, fusing the semantic priors of language with the spatial awareness of 3D visual features to guide the physical reasoning process.

Specifically, the textual description is first processed by a frozen OpenCLIP text encoder to extract a dense semantic embedding, denoted as $e_{text}$. Note that since both the 3D visual voxels and the textual embeddings are derived from the OpenCLIP latent space, they are inherently aligned, which avoids the need for complex cross-modal projection or pre-processing modules.
To achieve the cross-modal fusion, we draw inspiration from Feature-wise Linear Modulation (FiLM)~\cite{perez2018film}, which demonstrates remarkable success in visual reasoning tasks. Rather than naively concatenating this embedding with visual features, $e_{text}$ acts as a direct and highly effective global condition to modulate the 3D OpenCLIP visual voxels at multiple scales.

In our design, $e_{text}$ is passed through dedicated Multi-Layer Perceptrons (MLP) to predict layer-specific modulation parameters: the scaling factor $\gamma$ and the shifting factor $\beta$. Let $F_i$ represent the intermediate 3D voxel feature map at the $i$-th layer of our U-Net. The FiLM layer applies a channel-wise affine transformation to modulate this feature map:
\begin{equation}
F_i^{\prime} = \gamma_i(e_{text}) \cdot F_i + \beta_i(e_{text}).
\end{equation}
By interleaving these FiLM layers throughout the 3D U-Net, the network learns to dynamically emphasize or suppress specific visual feature channels based on linguistic cues.
This multi-scale, text-driven modulation ensures that the final extracted features are highly attuned to the physical attributes specified in the text, thereby enabling the accurate regression of the target physical parameters. We present a detailed comparison between our design and other multi-modal fusion modules in the \supp.

The physics loss $\mathcal{L}_{phys}$ is computed by enforcing supervision exclusively on the occupied voxels $\mathcal{G}$. Let $\mathcal{K} = \{E, \nu, \rho\}$ denote the set of target physical properties, corresponding to Young's modulus, Poisson's ratio, and density, respectively. We formulate the total loss as:
\begin{equation}
\mathcal{L}_{phys} = \frac{1}{N_{occ}} \sum_{p \in \mathcal{G}} \sum_{k \in \mathcal{K}} \| \hat{k}(p) - k^{GT}(p) \|^2,
\end{equation}
where $N_{occ}$ represents the total count of valid voxels, and $\hat{k}(p)$, $k^{GT}(p)$ denote the predicted and ground-truth physical parameters at voxel $p$.

\subsection{Dataset Generation}
\label{sec:data}
Unlike Pixie ~\cite{le2025pixiefastgeneralizablesupervised} and VoMP ~\cite{dagli2025vomppredictingvolumetricmechanical}, which supervise the prediction of each object by a single deterministic set of physical parameters, our approach resolves the inherent ambiguity of prediction by introducing text description as conditioning signals. This formulation supports downstream applications that require flexible object physics, \eg, increasing the robustness of robot manipulation policy by varying the potential plausible physics for the same object during training. However, such text-conditioned training data are currently unavailable at scale, motivating our automated pipeline for large-scale data generation.

\paragraph{Semantic Part Segmentation.}
To achieve precise text-driven control, \eg, accurately specifying whether a tree trunk is hardwood or softwood, or adjusting the stiffness of a car's tires, it is required that our dataset provide accurate part-level segmentation. To this end, we first extract 3D OpenCLIP voxels for each 3D object as in \cref{sec:get3dfeat}. OpenCLIP voxels have been successfully and widely adopted in object grounding and reasoning~\cite{shen2023F3RM}, and their text-aligned semantic features enable segmentation of objects based on corresponding text keywords.

To mitigate the ambiguity associated with single-view observations, we uniformly render 15 images from the upper hemisphere of each object. We then prompt \ac{vlm}~\cite{comanici2025gemini} to generate a text list encompassing all constituent parts it identifies for the object. Upon obtaining these text keywords, we compute the segmentation on the OpenCLIP voxels based on the cosine similarity between the text-visual features. To ensure robustness, we repeatedly generate five distinct sets of keywords per object and render their corresponding 3D segmentations, as illustrated in ~\cref{fig:extract3d}. The \ac{vlm} finally selects the highest-quality segmentation from these five candidates, which effectively prevents incorrect segmentations caused by the use of uncommon vocabulary for part names.

\paragraph{Text-Guided Property Assignment.}
After part-level segmentation, the next step is to generate the physical properties paired with text descriptions. To resolve the physical ambiguity from pure visual inputs, our guideline is to generate text first to guide the property assignment.
For each object, we randomly select a subset of parts and prompt \ac{vlm}~\cite{comanici2025gemini} to generate physically diverse but plausible textual descriptions, which act as the guideline for parameter assignment. To ground these assignments and prevent potential \ac{vlm} hallucination, for each description, we prompt \ac{vlm} to query the corresponding physics parameter range from online databases, \ie, Matweb~\cite{matweb}, Wikipedia~\cite{wikipedia}, and The Engineering Toolbox~\cite{engineeringtoolbox}. Once the physics ranges are acquired, we leverage \ac{vlm} to sample a corresponding parameter in the valid range. For instance, the description ``This is mahogany that might sink to the bottom,'' would lead to a density value near the upper end of the wood's density range ($800 \text{ kg/m}^3$ to $1300 \text{ kg/m}^3$), whereas ``This is softwood that can float on water'' yields a lower density. When an object comprises multiple parts, any unmentioned parts are assigned the mean value of their valid physical ranges.
 In addition, to ensure robust physical parameter assignment in the absence of detailed textual guidance, for each object, we
  include a simple default text description ``This is a [object name],'' where all parts are assigned the mean value in their
  valid ranges.
  We further filter generated assignments through simulation-based consistency checks: for each selected object and sampled
  physical setting, we generate a 100-frame simulation and reject cases with clear artifacts, poor segmentation, implausible
  property assignments, or behavior inconsistent with the generated description. The \ac{vlm} may resample the assignment up to
  five times; objects that still fail this automatic check are discarded.

\section{Method}
\label{sec:method}
\begin{figure}[t!]
  \centering
  \vspace{-3mm}
  \includegraphics[width=\linewidth]{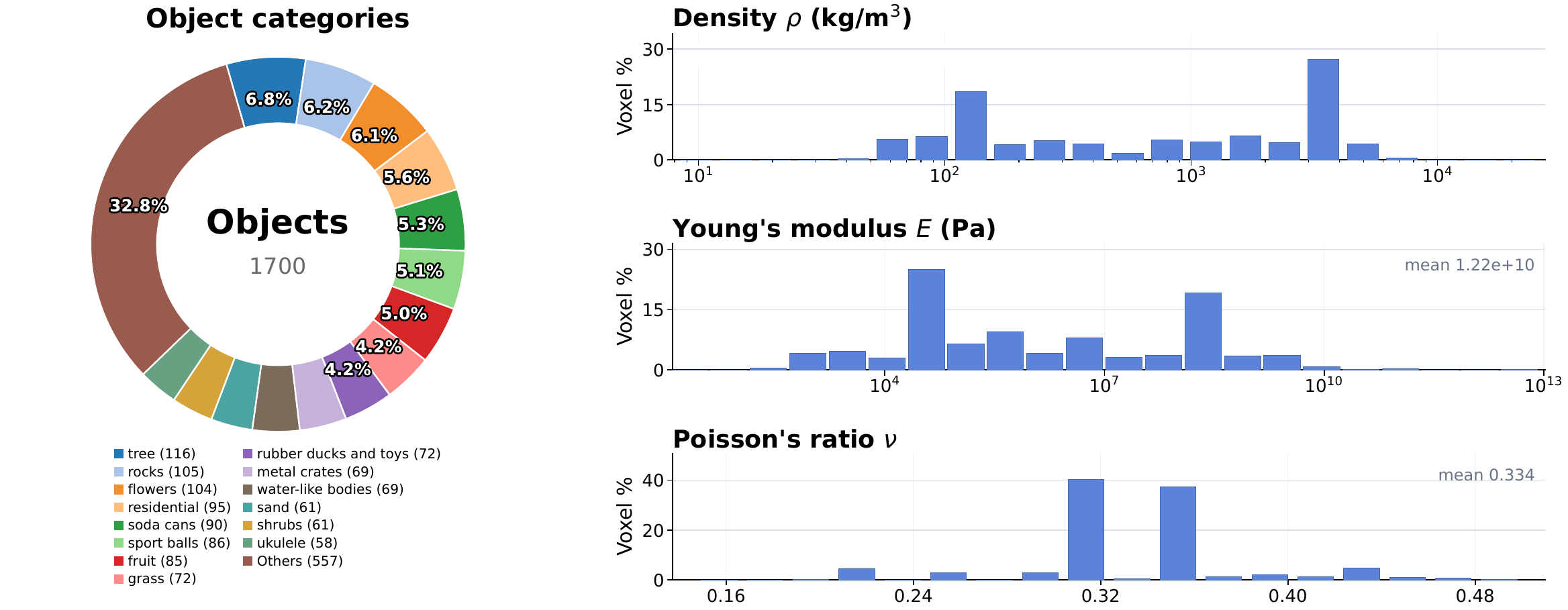}
  \caption{\textbf{Distribution of object categories and physical properties in \dataset.} \dataset contains 1,700 assets with multiple text-conditioned physical variants, covering diverse object categories as well as distributions of density, Young’s modulus, and Poisson’s ratio.}
  \label{fig:dataset_statistics}
  \vspace{-4mm}
\end{figure}

 To build a diverse and comprehensive dataset, we acquire our 3D assets from two distinct sources. First, we collect
  approximately 1,000 high-quality 3D meshes from the Objaverse dataset. Second, to capture the complexity of real-world
  objects, we gather diverse 2D images via Google Search and utilize SAM 3D Objects~\cite{sam3dteam2025sam3d3dfyimages} to
  reconstruct approximately 700 3D Gaussian Splatting (\ac{gs}) models. This yields a total base of 1,700 unique 3D assets. As
  detailed above, we generate 8 distinct textual descriptions and their corresponding physical parameters for each object.
  Consequently, our dataset, \dataset, contains $1700 \times 8 = 13600$ data triplets, with each triplet comprising a 3D object
  representation, a text description, and the ground-truth physical parameters. We partition this collection at the object
  level into 1,575 training objects and 125 testing objects, corresponding to 12,600 training samples and 1,000 testing
  samples. The resulting category distribution is shown in Fig.~\ref{fig:dataset_statistics}, and additional dataset galleries
  are included in \supp.

\section{Experiment}

We conduct a comprehensive evaluation of \modelname to validate its ability to recover accurate physical constants from visual and linguistic inputs. Our experiments focus on two primary objectives: (1) \textbf{Quantitative Analysis}, where we measure the precision of predicted physics properties on our dataset \dataset and a real-world dataset ABO-500; and (2) \textbf{Qualitative Simulation}, where we demonstrate the practical utility of our framework by applying the predicted physics properties to realistic physics-based simulations, containing both \acl{gs} and mesh. Through these evaluations, we show that \modelname effectively bridges the gap between high-level semantic descriptions and low-level physical dynamics in an end-to-end manner.

\paragraph{Dataset and Settings.}
We evaluate text-conditioned physical parameter prediction on the test set of \dataset, detailed in \cref{sec:data}. We measure the discrepancy between the estimated and ground-truth physical parameters. Furthermore, we assess the visual and dynamic quality of renderings from simulations driven by the predicted properties and the ground-truth.

 Furthermore, to rigorously assess real-world generalization, we incorporate the ABO-500 dataset as an additional out-of-
  distribution test set. The ABO-500 dataset contains the density and mass of 500 real-world objects, which allows us to
  evaluate the discrepancy between our predicted physical properties and the real-world ground truth. Since this dataset lacks
  detailed physical descriptions, we solely use text prompts following the simple template: ``This is a [object name].''
  We also include additional human-annotated and cross-dataset evaluations in \supp, where natural language descriptions are
  provided by annotators after observing simulated object behavior rather than generated by our data pipeline.

\begin{table}[t!]
  \centering
  \caption{\textbf{Quantitative comparison on \dataset.} We evaluate both rendering quality (PSNR, SSIM, LPIPS) and physical property prediction accuracy (Avg. Phys. err, $\log E$ err, $\nu$ err, $\log \rho$ err). ``$-$'' indicates that the method cannot predict the corresponding physical parameter. \textbf{Bold} and \underline{underline} denote the best and second-best results, respectively. Our method, particularly when equipped with OpenCLIP, significantly outperforms all baselines across both rendering and physics metrics.}
  \label{tab:results}

  \resizebox{\textwidth}{!}{%
    \begin{tabular}{l ccc cccc}
      \toprule
      Method & PSNR $\uparrow$ & SSIM $\uparrow$ & LPIPS $\downarrow$ & Avg. Phys. err $\downarrow$ & $\log E$ err $\downarrow$ & $\nu$ err $\downarrow$ & $\log \rho$ err $\downarrow$ \\
      \midrule
      \multicolumn{8}{l}{\textit{DreamPhysics}} \\
      \quad 1 epoch & 19.412 & 0.881 & 0.212 & $-$ & 2.351 & $-$ & $-$ \\
      \quad 25 epochs & 19.515 & 0.882 & 0.185 & $-$ & 1.405 & $-$ & $-$ \\
      \quad 50 epochs & 19.223 & 0.881 & 0.178 & $-$ & 1.362 & $-$ & $-$ \\
      \midrule
      \multicolumn{8}{l}{\textit{OmniPhysGS }} \\
      \quad 1 epoch & 17.955 & 0.883 & 0.325 & $-$ & $-$ & $-$ & $-$ \\
      \quad 2 epochs & 17.912 & 0.883 & 0.310 & $-$ & $-$ & $-$ & $-$ \\
      \quad 5 epochs & 17.880 & 0.884 & 0.305 & $-$ & $-$ & $-$ & $-$ \\
      \midrule
      \textit{NeRF2Physics} & 18.554 & 0.889 & 0.245 & 0.841 & 0.585 & 0.455 & 0.982 \\
      \midrule
      \textit{Gemini} (avg.) & 21.124 & 0.869 & 0.205 & 0.218 & 0.227 & 0.235 & 0.192 \\
      \midrule
      \multicolumn{8}{l}{\textit{Pixie} (uncond. eval only)} \\
      \quad Occupancy & 17.901 & 0.868 & 0.285 & 0.122 & 0.145 & 0.121 & 0.101 \\
      \quad RGB & 18.695 & 0.863 & 0.262 & 0.121 & 0.191 & 0.075 & 0.097 \\
      \quad Clip & 23.421 & 0.908 & 0.092 & 0.078 & 0.062 & 0.065 & 0.108 \\
      \midrule
      \textbf{Ours (w/ Clip)} & \underline{25.105} & \underline{0.916} & \underline{0.091} & \underline{0.069} & \underline{0.058} & \underline{0.061} & \underline{0.089} \\
      \textbf{Ours (w/ OpenClip)} & \textbf{26.012} & \textbf{0.921} & \textbf{0.089} & \textbf{0.059} & \textbf{0.042} & \textbf{0.051} & \textbf{0.083} \\
      \bottomrule
    \end{tabular}%
  }
\end{table}
\paragraph{Implementation Detail.} \modelname is trained using the Adam optimizer with the initial learning rate $10^{-4}$ on 2 NVIDIA RTX A800 GPUs. Training takes 2 days with a batch size of 4 per GPU and 100 epochs. Training hyperparameters are selected once using a small 3,000/300 train/test split drawn from the training objects and then kept fixed for all reported experiments. For both training and metric computation, we apply a log transform to $E$ and $\rho$, and normalize all $\log E$, $\nu$, and $\log \rho$ values to the range $[-1, 1]$. The \supp contains the model architecture and OpenCLIP version.

\paragraph{Baselines.}
We compare \modelname against several representative methods. Both OmniPhysGS~\cite{lin2025omniphysgs} and DreamPhysics~\cite{huang2025dreamphysics} utilize video diffusion models to optimize the physics simulation process online.
NeRF2Physics~\cite{zhai2024physical} uses GPT~\cite{achiam2023gpt} to generate material text prompts and their physical properties, mapping them to 3D points via text-image CLIP similarity.
We also test representative \ac{vlm}, Gemini, which we ask to predict the part name and corresponding physics property, and then assign them back to the object corresponding to OpenCLIP feature similarity. To alleviate the randomness of \ac{vlm} output, we conduct 5 queries and take the average as the final result of Gemini.
Pixie~\cite{le2025pixiefastgeneralizablesupervised} is a recent feed-forward physics generation method that relies solely on visual features. Since the original dataset for Pixie is not publicly available, we retrain it on our dataset to ensure a fair comparison. Furthermore, as Pixie does not support textual descriptions as input, we evaluate it exclusively under the default unconditional setting described in \cref{sec:data}. We additionally report unconditional cross-dataset comparisons with VoMP~\cite{dagli2025vomppredictingvolumetricmechanical} in \supp.

\paragraph{Evaluation Metrics.}
In \dataset, we evaluate using physical and visual metrics. For physical accuracy, we compute the Mean Squared Error (MSE) for Young's modulus $E$, Poisson's ratio $\nu$, and density $\rho$. To handle large magnitude variations, $E$ and $\rho$ are evaluated in log-space ($\log_{10}$). This is especially important for Young's modulus, which spans many orders of magnitude across materials; raw absolute error would be dominated by stiff materials and would underweight soft materials where small absolute changes can strongly alter dynamics. For visual fidelity, we simulate and render images with the predicted properties using PhysGaussian~\cite{xie2024physgaussian}
for \acl{gs}, and Isaac Sim~\cite{makoviychuk2021isaacgymhighperformance} for meshes. The rendered frames are then evaluated against ground truth using Peak Signal-to-Noise Ratio (PSNR), Structural Similarity Index Measure (SSIM), and Learned Perceptual Image Patch Similarity (LPIPS).

On the real-world ABO-500 mass dataset, we report four metrics: ADE (Average Displacement Error), ARE (Average Relative Error), ALDE (Average Log Displacement Error), and MnRE (Minimum Ratio Error). Specifically, ADE measures the absolute numerical difference in weight, while ARE captures the percentage of this error relative to the object's true weight, which is crucial for lighter items. Furthermore, ALDE evaluates the error in logarithmic space. MnRE avoids bias toward systematic over- or underestimation and reduces sensitivity to outliers, making it particularly effective for evaluating physical quantity predictions across heterogeneous samples.

\subsubsection{Quantitative Result}
\begin{figure}[t!]
    \centering

    \begin{minipage}[c]{0.58\textwidth}
        \centering
        \captionof{table}{\textbf{Quantitative comparison on ABO-500 dataset.}
        ALDE: average log displacement error; ADE: average displacement error;
        ARE: average relative error; MnRE: minimum ratio error
        \begin{math}\left(\frac{1}{N}\sum_{i=1}^{N}\min\!\left(\frac{y_i}{\hat{y}_i},\frac{\hat{y}_i}{y_i}\right)\right)\end{math}.
        Lower is better for ALDE/ADE/ARE; higher is better for MnRE.}
        \label{tab:metrics}

        \vspace{2mm} %

        \resizebox{\linewidth}{!}{
            \renewcommand{\arraystretch}{1.05}
            \begin{tabular}{l c c c c}
                \toprule
                Method & ALDE (\begin{math}\downarrow\end{math}) & ADE (\begin{math}\downarrow\end{math}) & ARE (\begin{math}\downarrow\end{math}) & MnRE (\begin{math}\uparrow\end{math}) \\
                \midrule
                NeRF2Physics & 0.732 & 11.826 & 0.989 & 0.573 \\
                Pixie        & 0.782 & 12.565 & 0.971 & 0.568 \\
                Ours         & \textbf{0.652} & \textbf{9.433} & \textbf{0.901} & \textbf{0.582} \\
                \bottomrule
            \end{tabular}
        }
    \end{minipage}\hfill %
    \begin{minipage}[c]{0.4\textwidth}
        \centering
        \includegraphics[width=\linewidth]{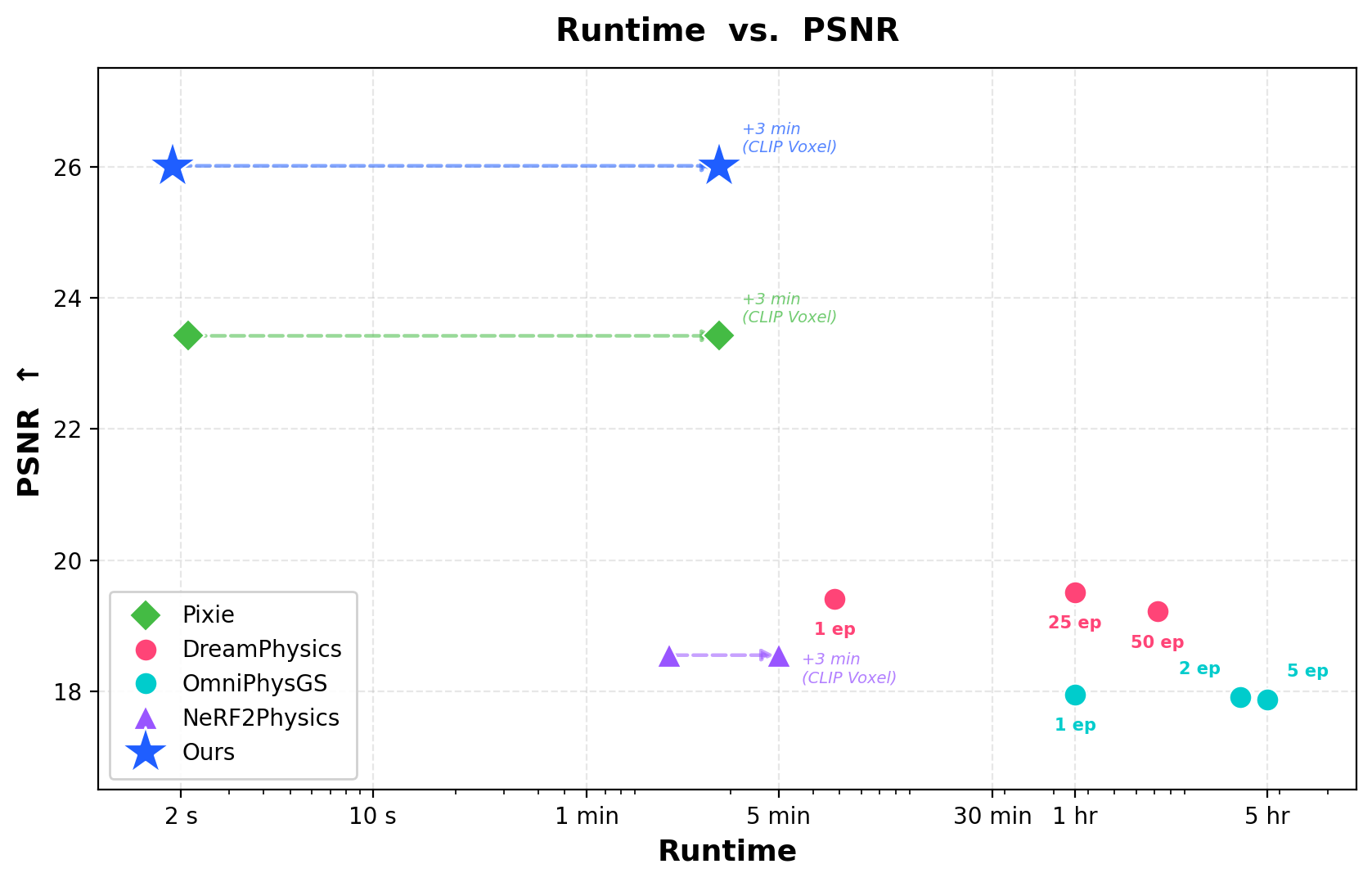}
        \captionof{figure}{Runtime vs. PSNR. Ours achieves the best PSNR (26.01) in under 2s of inference, orders of magnitude faster than optimization-based methods.}
        \label{fig:vlm}
    \end{minipage}

\end{figure}

As shown in \cref{tab:results}, our method exhibits a significant performance margin over existing baselines. \modelname with
  the OpenCLIP features achieves a PSNR of 26.012, outperforming the state-of-the-art Pixie by approximately 2.6 dB. This
  significant improvement is primarily attributed to our cross-modal modulation utilizing the text conditions, coupled with
  OpenCLIP's richer semantic representations compared to standard CLIP; they together empower our model to make finer-grained
  physical predictions. Pixie, while making reasonable predictions, is constrained to the default physical conditions because
  it solely takes CLIP voxel features as input. This reliance on purely visual input suffers from inherent visual ambiguity,
  which is more evident on the ABO-500 object mass estimation benchmark.
As shown in~\cref{tab:metrics}, our method consistently improves accuracy over NeRF2Physics and Pixie, achieving lower ALDE, ADE, ARE, and higher MnRE, which suggests that \modelname not only achieves a good performance on text-conditioned prediction on our dataset, but also generalizes well in the out-of-distribution real-world dataset across diverse physical scenarios.

 Since NeRF2Physics directly relies on GPT for physical property predictions, it inherits inaccurate physical priors and
  \ac{vlm} hallucinations. As a result, its performance in terms of physical accuracy and visual fidelity is relatively low.
  Similar to NeRF2Physics, directly leveraging Gemini as an agent that predicts physics properties cannot achieve high
  precision. While
  \dataset also employs Gemini for part segmentation and physical property retrieval during data generation, we mitigate these
  limitations through a multiple-sampling strategy for robust part segmentation and text-guided property assignment.
  Furthermore, we ground our queries in authoritative online databases rather than relying solely on the VLM's internal
  knowledge for physical properties. This also highlights a key \textit{asymmetry}: \ac{vlm}s can generate coherent part
  descriptions and then produce compatible physical parameters in a forward manner, but they struggle with the inverse problem
  that requires specific property estimation from a given text description, because their internal physics knowledge is not
  well calibrated for accurate, instance-specific parameter estimation. Consequently, although we utilize Gemini for data
  generation, our method leverages large-scale supervised training to improve generalizability and substantially outperform
  pure VLM-based baselines across all evaluation metrics.

Since NeRF2Physics directly relies on GPT for physical property predictions, it inherits inaccurate physical priors and \ac{vlm} hallucinations. As a result, its performance in terms of physical accuracy and visual fidelity is relatively low. Similar to NeRF2Physics, directly leveraging Gemini as an agent that predicts physics properties cannot achieve high precision. While
\dataset also employs Gemini for part segmentation and physical property retrieval during data generation, we mitigate these limitations through a multiple-sampling strategy for robust part segmentation and text-guided property assignment. Furthermore, we ground our queries in authoritative online databases rather than relying solely on the VLM's internal knowledge for physical properties. This also highlights a key \textit{asymmetry}: \ac{vlm}s can generate coherent part descriptions and then produce compatible physical parameters in a forward manner, but they struggle with the inverse problem that requires specific property estimation from a given text description, because their internal physics knowledge is not calibrated for accurate, instance-specific parameter estimation. Consequently, although we utilize Gemini for data generation, our method leverages large-scale training to improve generalizability and substantially outperform pure VLM-based baselines across all evaluation metrics.

\subsubsection{Runtime.}
\cref{fig:vlm} plots PSNR versus runtime (log-scale). Our method attains the highest PSNR
(26.01) with 2s inference, while optimization-based baselines (DreamPhysics and OmniPhysGS)
require minutes to hours, depending on the number of epochs, and yield lower PSNR.
For Pixie, NeRF2Physics, and our method, we additionally show a second point that accounts for
approximately 3 minutes of CLIP voxel feature extraction as input preprocessing.

\begin{figure}[t!]
    \centering
    \includegraphics[width=\linewidth]{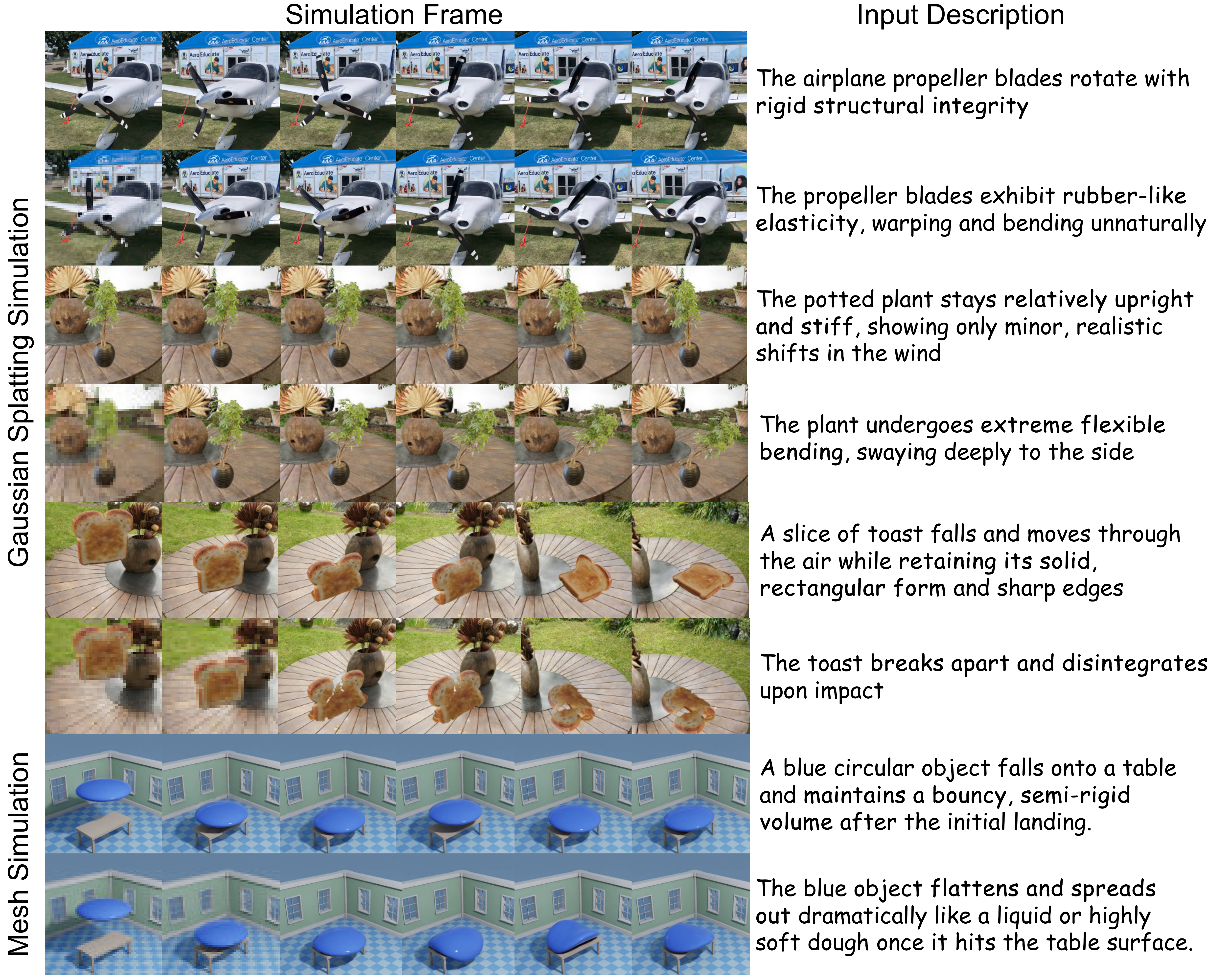}
    \caption{\textbf{Qualitative verification of physical parameter prediction.} To evaluate the effectiveness of our parameter estimation, we visualize simulations under identical boundary conditions but with different predicted physical properties. Each pair of rows demonstrates how various descriptions alter the simulation behavior: Rows 1--2 for rigid vs. elastic propeller blades, Rows 3--4 for stiff vs. flexible plant stems, Rows 5--6 for structural integrity vs. brittle fracture of toast, and Rows 7--8 for semi-rigid vs. large deformation of a soft body.}
    \label{fig:qualitative_results}
\end{figure}
\FloatBarrier
\subsubsection{Qualitative Results}

As illustrated in Fig.~\ref{fig:qualitative_results}, the versatility of our voxelized prediction approach allows the predicted physical properties to be integrated into diverse simulation pipelines.
For the first six rows, we employ the \ac{mpm} to drive the dynamic behavior of \acl{gs} representations, capturing complex physical effects under varying material properties: from rigid rotation to elastic warping of the propeller (Rows 1-2), from gentle swaying to pronounced bending of the ficus in the wind (Rows 3-4), and from an intact collision to brittle fracture of the falling bread (Rows 5-6).

Furthermore, to demonstrate the cross-representation capability, the final two rows present deformable object simulations conducted in Isaac Sim using the \ac{fem}. Despite the change in both the physical solver and the 3D  representation, our model consistently predicts accurate parameters that govern realistic material behaviors, such as volume-preserving bouncing versus large deformation. This highlights that our method learns intrinsic physical attributes that are independent of specific simulators or 3D representations.

\begin{wrapfigure}[18]{r}{0.52\linewidth}
    \centering
    \vspace{-8mm}
    \includegraphics[width=\linewidth]{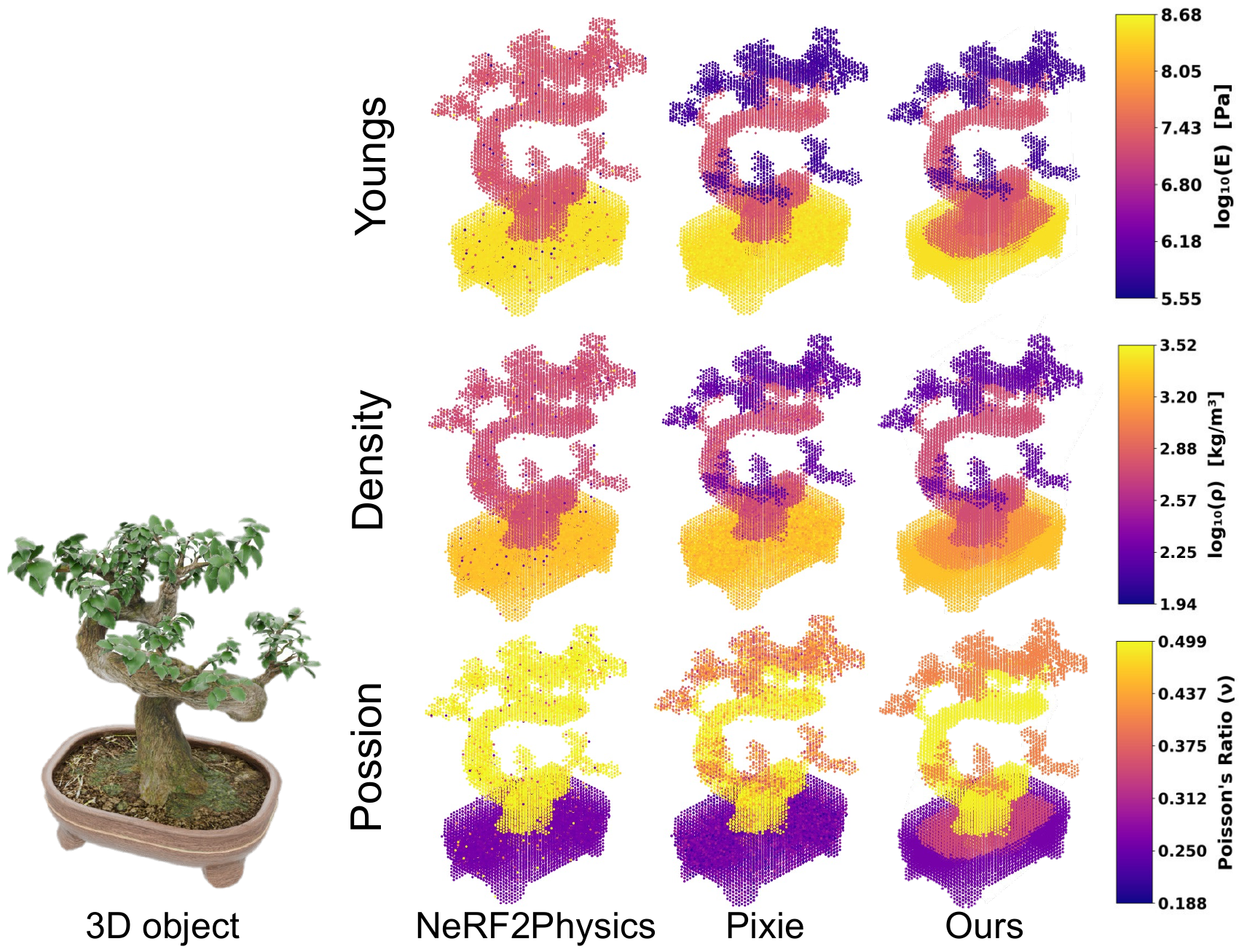}
    \caption{\textbf{Voxel prediction across different methods.} Compared to baselines that struggle with structural granularity (NeRF2Physics) or material boundaries (Pixie), our framework achieves highly precise material decomposition.}
    \label{fig:heatmap}
    \vspace{-2mm}
\end{wrapfigure}

 As illustrated in Fig.~\ref{fig:heatmap}, our voxel-based prediction framework achieves significantly higher precision in
  material decomposition compared to existing methods. A key limitation of NeRF2Physics is its lack of structural granularity,
  which predicts nearly identical physical attributes for different parts, such as assigning uniform Young's modulus values to
  both the foliage and the trunk. On the other hand, Pixie struggles with material boundaries at the base of the object,
  predicting that the soil and the ceramic pot have the same physical properties. In contrast, our method successfully recovers
  the sharp discontinuities between distinct materials. It accurately distinguishes the rigid trunk from the soft leaf clusters
  and maintains a clear boundary between the soil and the rigid container. This heatmap confirms that our model captures
  intrinsic semantic-physical correlations beyond simple global averages.

\section{Conclusion and Future Work}
\label{sec:conclusion}
 We presented \modelname, a text-conditioned feed-forward framework that automates dense physical property assignment for
  diverse 3D assets. Our method uses a unified voxel field representation to bridge high-level linguistic semantics with low-
  level physical dynamics in a scalable manner. To support this cross-modal paradigm, we designed an automated, \ac{vlm}-driven
  generation pipeline and constructed \dataset, a large-scale, multi-part 3D physics dataset for training and testing across
  diverse object categories.

  Extensive experiments validate \modelname. Quantitatively, it achieves state-of-the-art accuracy on our dataset and the real-world ABO-500 benchmark. Qualitatively, our method integrates into simulation pipelines, including
  \ac{mpm} and \ac{fem}, across objects such as a bonsai, a propeller, and bread. These results demonstrate its potential for larger, complex interactive environments. In the future, we plan to extend our framework by incorporating
  additional physical properties, such as surface friction and damping coefficients, and by exploring higher-resolution fields for objects with very fine structures.

\subsubsection*{Acknowledgements}
This work was supported in part by National Natural Science Foundation of China under Grant 62272011.

\clearpage
\bibliographystyle{splncs04}
\bibliography{reference}
\clearpage
\appendix

\setcounter{page}{1}
\renewcommand{\thefigure}{S.\arabic{figure}}
\renewcommand{\thetable}{S.\arabic{table}}
\renewcommand{\theequation}{S.\arabic{equation}}

\renewcommand\thefigure{A\arabic{figure}}
\setcounter{figure}{0}
\renewcommand\thetable{A\arabic{table}}
\setcounter{table}{0}
\renewcommand\theequation{A\arabic{equation}}
\setcounter{equation}{0}
\setcounter{footnote}{0}

\section{\dataset Dataset Details}
\label{sec:dataset_details}

\subsection{Overview}
As introduced in the main text, \dataset is a large-scale, text-conditioned 3D physical property dataset designed to resolve the inherent ambiguity of predicting object mechanics from pure visual inputs. The dataset is built upon a diverse collection of 1,700 unique 3D assets, which are sourced from two distinct domains to ensure both quality and real-world complexity: approximately 1,000 high-quality 3D meshes from the Objaverse dataset, and 700 3D \acf{gs} models reconstructed from in-the-wild 2D images using SAM 3D Objects.

To support flexible, text-driven physical predictions, we generate 8 distinct textual descriptions paired with corresponding physical parameters for each base object. Consequently, \dataset comprises a total of 13,600 data triplets. Each triplet consists of:
\begin{enumerate}
    \item A 3D object representation (dense OpenCLIP semantic voxels extracted from multi-view renderings).
    \item A text description specifying the material or physical state of the object's constituent parts.
    \item The ground-truth dense physical parameters, specifically Young's Modulus ($E$), Poisson's Ratio ($\nu$), and Density ($\rho$), assigned at the voxel level.
\end{enumerate}
The entire dataset is partitioned at the object level into a training set of 12,600 samples from 1,575 objects and a testing set of 1,000 samples from 125 held-out objects, providing a robust benchmark for text-conditioned 3D physical parameter regression.

\subsection{Gallery with Physics Property}

\begin{figure}[t!]
  \centering
  \includegraphics[width=0.99\textwidth]{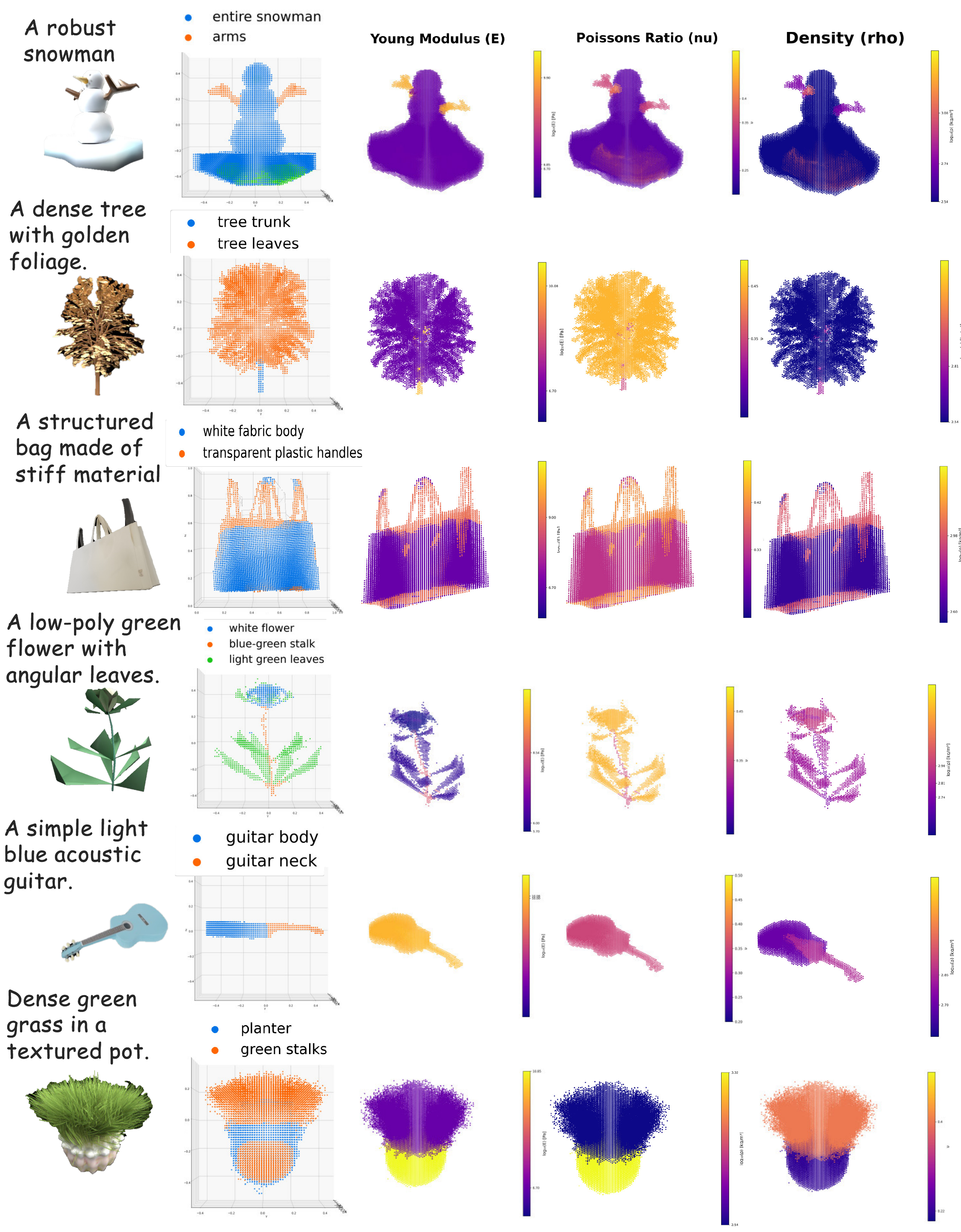}
  \caption{\textbf{Gallery of \dataset.} This figure illustrates the fine-grained, text-conditioned physical annotations in our dataset. From left to right: the original 3D object rendering, the part-level semantic segmentation (extracted via OpenCLIP and \ac{vlm}), and the corresponding voxel-wise physical property heatmaps for Young's Modulus ($E$), Poisson's Ratio ($\nu$), and Density ($\rho$). Notice how distinct semantic parts within the same object (\eg, the tree trunk versus leaves, or the bag's body versus handles) are assigned accurately localized and physically plausible parameters.}
  \label{fig:dataset_example}
\end{figure}

\begin{figure}[t!]
  \centering
  \includegraphics[width=0.99\textwidth]{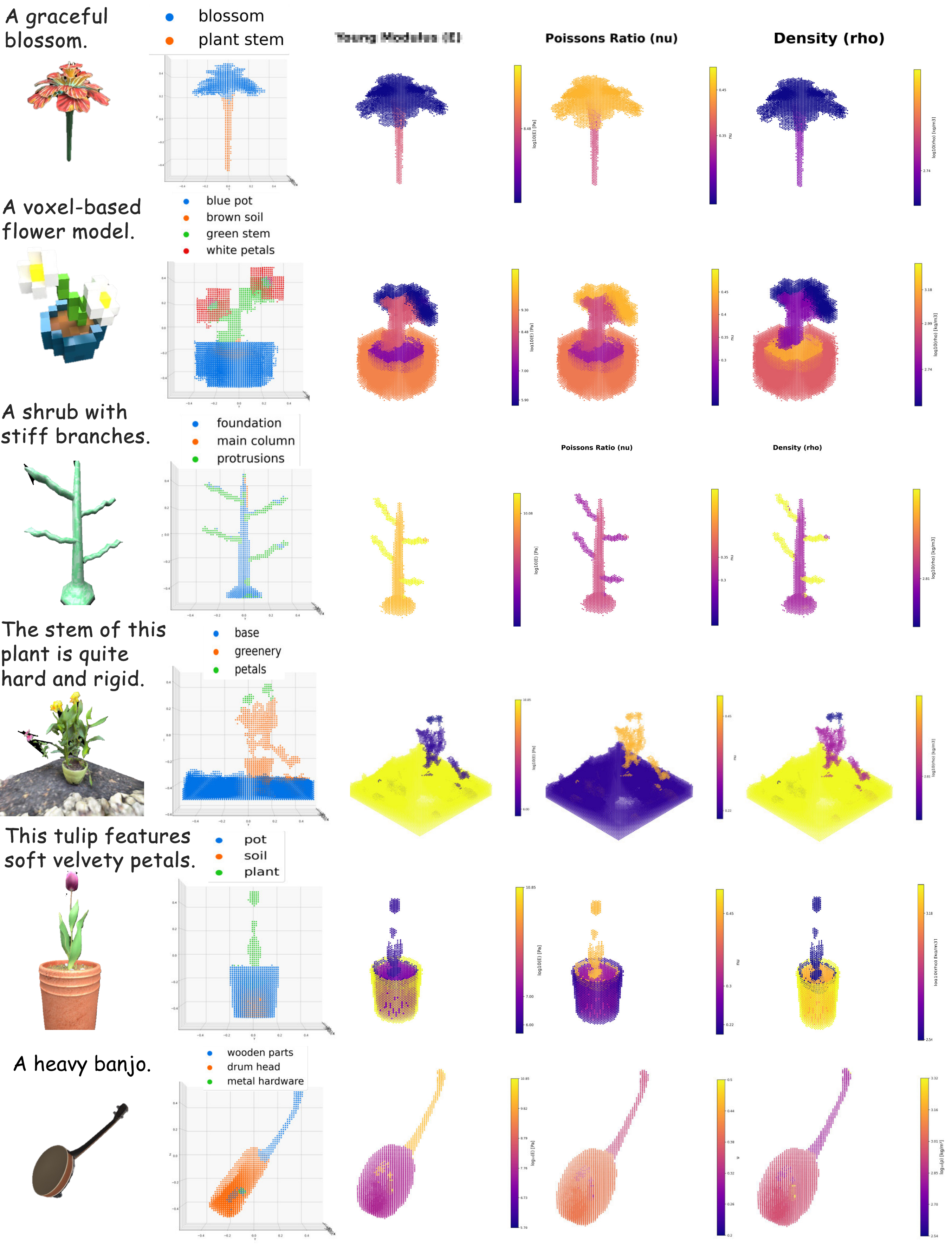}
  \caption{\textbf{Gallery of \dataset.} }
  \label{fig:dataset_example2}
\end{figure}

To intuitively demonstrate the diversity and granularity of \dataset, Figure \ref{fig:dataset_example} provides a visual gallery showcasing various objects alongside their part-level segmentations and corresponding physical property heatmaps.

\paragraph{Multi-View Renderings and Semantic Segmentation.}
To mitigate the ambiguity of single-view observations and extract robust 3D OpenCLIP voxels, we uniformly render multiple views from the upper hemisphere of each object. Based on these visual features and \ac{vlm}-generated keywords, objects are accurately segmented into constituent parts. As shown in the second column of Figure \ref{fig:dataset_example}, a snowman is segmented into its body, arms, and nose; a potted grass is separated into the planter and the green grass; and a bag is distinctly divided into the fabric body and plastic handles. This fine-grained part-level segmentation is crucial for applying localized, text-driven physical controls.

\paragraph{Voxel-Wise Physical Property Heatmaps.}
Building upon the part segmentation, \dataset provides dense, voxel-wise annotations for three key physical properties. As visualized in the rightmost three columns of Figure \ref{fig:dataset_example}, different semantic parts within the same object exhibit distinctly different physical values, reflecting their assigned materials. For example, the wooden trunk of the tree shows a significantly higher Young's Modulus ($E$) compared to its leaves, while the plastic handles of the bag exhibit different densities ($\rho$) and elasticity compared to the fabric body. These detailed heatmaps confirm that our automated, text-guided pipeline successfully assigns physically plausible and spatially accurate parameters across a wide variety of complex 3D objects.

\begin{figure}[t!]
  \centering
  \includegraphics[width=0.99\textwidth]{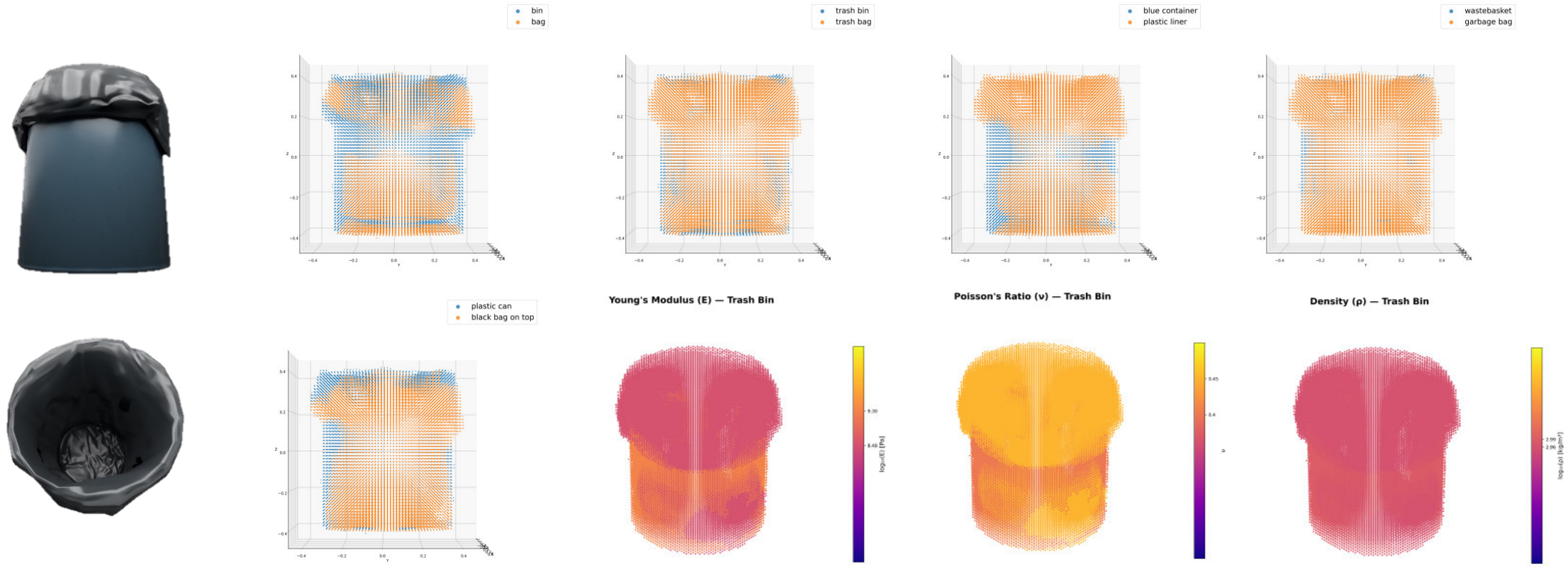}
  \caption{\textbf{Failure case in segmentation.} Due to the tight adherence between the trash can's outer shell and the inner plastic bag, most voxels encapsulate both components post-voxelization. This renders perfect segmentation unattainable regardless of the part name queried in the \ac{vlm}, since the features within any given voxel are effectively an average of those from both the plastic bag and the trash bin.}
  \label{fig:segmentation_failure}
\end{figure}
\subsection{OpenCLIP-Driven 3D Semantic Segmentation}
\label{sec:openclip_segmentation}

\paragraph{Overview}
To endow our volumetric representation with open-vocabulary semantic understanding suitable for physics-based simulation, we distill dense, high-dimensional semantic features from a pre-trained 2D vision-language foundation model (specifically OpenCLIP) into a continuous 3D feature field. Furthermore, to ensure physical accuracy during stress propagation, we extend these surface features into the unobservable internal volume using a robust ray-casting strategy.

\subsubsection{2D Feature Extraction}
Given a set of multi-view RGB images $ \{ \mathbf{I}_k \} $ of the scene, we first extract 2D semantic feature maps using a pre-trained OpenCLIP image encoder. Since vision transformers typically output features at a lower resolution than the input image, we obtain a dense feature map $ \mathbf{F}_k \in \mathbb{R}^{512 \times 512 \times 768} $ for each image $ \mathbf{I}_k $, where $ 768 $ is the dimensionality of the OpenCLIP latent space.

\subsubsection{3D Feature Distillation via Volumetric Rendering}
Unlike heuristic multi-view projection methods that suffer from occlusion ambiguities, we lift the 2D semantic features into 3D using a physics-inspired volumetric rendering framework. We freeze the density field $ \sigma $ obtained from a pre-trained RGB NeRF and introduce a continuous 3D feature field $ \Phi: \mathbb{R}^3 \to \mathbb{R}^D $, parameterized by a neural network or a feature grid.

During optimization, we render the expected 3D feature $ \hat{F}(\mathbf{r}) $ along a camera ray $ \mathbf{r} $ using the accumulated transmittance $ T(t) $. This naturally handles occlusions by dropping the transmittance to zero behind visible surfaces. We optimize the feature field $ \Phi $ by minimizing the discrepancy between the rendered ray features $ \hat{F}(\mathbf{r}) $ and the corresponding ground-truth OpenCLIP features $ F_k(\mathbf{r}) $ sampled from the 2D feature maps. We employ the negative cosine similarity loss as our distillation objective:
$$ \mathcal{L}_{\text{distill}} = - \frac{1}{N} \sum_{k} \sum_{\mathbf{r}} \frac{\hat{F}(\mathbf{r}) \cdot F_k(\mathbf{r})}{\| \hat{F}(\mathbf{r}) \|_2 \| F_k(\mathbf{r}) \|_2} $$
where $ N $ is the total number of sampled rays. By minimizing this loss, the 3D feature field learns to encode view-consistent semantic priors strictly on the visible surfaces.

\subsubsection{Internal Volume Filling via Even-Odd Rule}
While the volumetric rendering successfully distills features onto the object's surface, physics-based simulations require semantic and physical properties (\eg, Young's modulus $ E $, Poisson's ratio $ \nu $) to be defined throughout the object's interior. Since internal voids are unobservable from external cameras, their features remain undefined.

To bridge this gap, we employ a hierarchical ray-casting strategy based on the classic even-odd rule. For a given unassigned voxel $ \mathbf{v} $ in the 3D space, we cast multiple rays uniformly distributed over a sphere towards infinity. Along each ray, we count the number of intersections with the object's surface (defined by a density threshold on $ \sigma $). If the intersection count is consistently odd across the majority of the casted rays, the voxel $ \mathbf{v} $ is robustly classified as being inside the object, effectively handling complex concavities.

Finally, for each identified internal voxel, we assign its semantic feature by performing nearest-neighbor interpolation from the optimized surface feature field $ \Phi $. This ensures that the internal volume inherits semantically and physically consistent properties from its closest bounding surface, yielding a complete volumetric representation ready for downstream simulation.

\subsubsection{Open-Vocabulary Querying}
Once the 3D feature distillation is complete, the scene can be queried using natural language. Given a user-provided text prompt (\eg, "a red apple" or "the handle of the mug"), we encode the text into a 768-dimensional embedding $\mathbf{t}$ using the OpenCLIP text encoder.

To perform 3D semantic segmentation, we compute the cosine similarity between the text embedding $\mathbf{t}$ and the distilled 3D feature $\mathbf{f}_p$ of each primitive:
$ S(p, \mathbf{t}) = \frac{\mathbf{f}_p \cdot \mathbf{t}}{\| \mathbf{f}_p \|_2 \| \mathbf{t} \|_2} $
Primitives with a similarity score $S(p, \mathbf{t})$ exceeding a predefined threshold are segmented as belonging to the target object. This continuous similarity field not only enables robust part-level segmentation but also facilitates downstream tasks such as language-guided robotic grasping and interaction.

\subsection{\ac{vlm} labeling}

\begin{figure}[t!]
  \centering
  \includegraphics[width=0.99\textwidth]{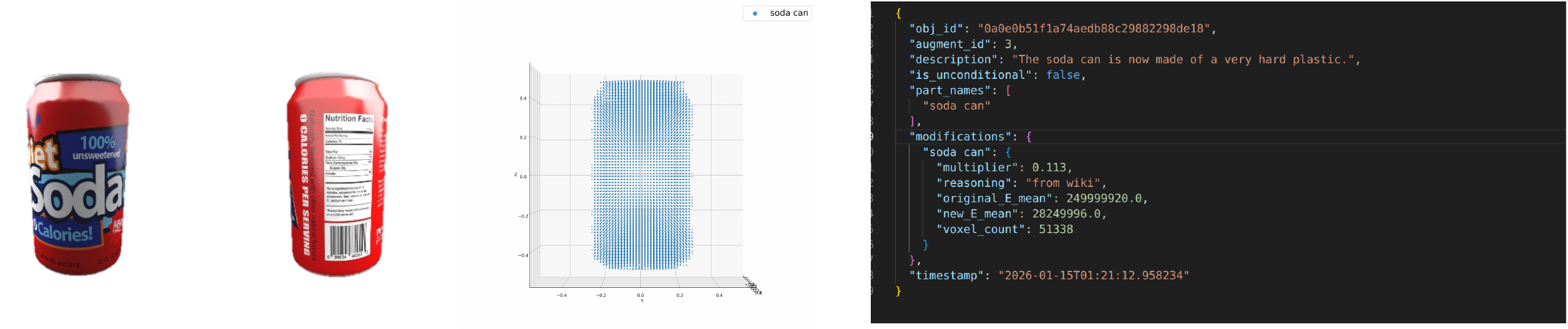}
  \caption{\textbf{Example of VLM-driven physical property augmentation.} The figure illustrates the automated data annotation process for a 3D object. From left to right: multi-view RGB images of a soda can, its corresponding 3D point cloud representation, and the VLM-generated JSON metadata. Based on the augmented semantic description (``The soda can is now made of a very hard plastic''), the VLM successfully modifies the object's physical parameters, such as scaling the Young's modulus ($E$) with a specific multiplier.}
  \label{fig:vlm_labeling}
\end{figure}

To further illustrate the automated data generation pipeline described in Section \ref{sec:data}, Figure \ref{fig:vlm_labeling} provides a concrete example of our text-guided property assignment process. As discussed in the main text, relying solely on visual inputs for physical parameter prediction is inherently ill-posed. To resolve this, we utilize a \ac{vlm} to generate diverse, physically plausible textual descriptions and assign corresponding physical parameters.

In this example, the input consists of multi-view RGB renderings of a soda can and its extracted 3D representation. During the augmentation phase, the \ac{vlm} generates a novel semantic condition: ``The soda can is now made of a very hard plastic.'' Rather than blindly guessing the physical values, the \ac{vlm} grounds its assignment by querying external knowledge bases. As shown in the generated JSON metadata, the model explicitly provides its reasoning trace (``from wiki'') to prevent hallucination.
\begin{table}[h]
\centering
\small
\caption{\textbf{Evaluation on the human-annotated test set.} The test descriptions are written by human annotators after watching simulated object behavior, providing a cleaner evaluation of text-conditioned generalization beyond the generated \dataset descriptions.}
\label{tab:human_annotated_eval}
\begin{tabular}{lcccc}
\toprule
\textbf{Model} & $\log E$ err. & $\nu$ err. & $\log \rho$ err. & $\rho$ err. \\
\midrule
PhysDreamer & 3.674 & 0.072 & 0.411 & 1107.99 \\
PhysX-Anything & 1.205 & 0.078 & 0.496 & 1399.63 \\
Pixie & 2.268 & 0.075 & 0.777 & 1268.79 \\
VoMP & 0.958 & 0.090 & 0.458 & 1330.10 \\
Ours w/o text & 0.624 & 0.079 & 0.287 & 603.47 \\
Ours & \textbf{0.181} & \textbf{0.056} & \textbf{0.102} & \textbf{296.63} \\
\bottomrule
\end{tabular}
\end{table}

Based on this grounded knowledge, the \ac{vlm} determines an appropriate scaling factor (multiplier: $0.113$) for the material transition from standard aluminum to hard plastic. This multiplier is then applied to the default Young's Modulus ($E$) of the object (``original\_E\_mean'': $249999920.0$) to compute the updated physical parameter (``new\_E\_mean'': $28249996.8$) for the specified part (``soda can''). This transparent, physics-aware reasoning process ensures that the generated text-parameter pairs in \dataset remain strictly consistent and physically accurate across all $13,600$ data triplets.

\section{Additional Evaluation and Analysis}
\subsection{Human-Annotated Text Generalization}
To further evaluate text generalization beyond \dataset, we construct a human-annotated test set of 30 objects. For each material category, we query The Engineering Toolbox to obtain plausible physical-property ranges, retrieve corresponding object images, and reconstruct 3D assets using SAM 3D Objects. We sample two material-property configurations from the queried ranges and run physics simulation for each sample. Human annotators then watch these videos and provide the corresponding natural language descriptions, making this evaluation independent of the text-generation pipeline used to build \dataset.

\subsection{Unconditional Cross-Dataset Validation}
To evaluate whether our framework is tied only to \dataset, we also compare Pixie, VoMP, and our model under an unconditional protocol using an empty text input. We train and test the methods on the Pixie dataset, the VoMP dataset, and \dataset, respectively. This setting is not designed to test text controllability; rather, it checks whether the voxel-based feed-forward architecture remains competitive when reduced to the same unconditional setting as prior feed-forward methods.

\begin{table}[h]
\centering
\caption{\textbf{Unconditional cross-dataset evaluation.} All methods are evaluated with empty text input for direct comparability with feed-forward baselines that do not support text conditioning.}
\label{tab:cross_dataset_uncond}
\begin{tabular}{llccc}
\toprule
\textbf{Model} & \textbf{Metric} & \textbf{Pixie dataset} & \textbf{VoMP dataset} & \textbf{\dataset} \\
\midrule
Pixie & $\log E$ err. & \textbf{0.022} & 0.330 & \underline{0.059} \\
      & $\nu$ err. & \underline{0.034} & 0.043 & 0.062 \\
      & $\rho$ err. & \underline{105.45} & 847.86 & \underline{153.72} \\
\midrule
VoMP & $\log E$ err. & 0.076 & \textbf{0.042} & 0.080 \\
     & $\nu$ err. & 0.048 & \underline{0.025} & \underline{0.058} \\
     & $\rho$ err. & 213.6 & \underline{113.31} & 262.78 \\
\midrule
Ours & $\log E$ err. & \underline{0.063} & \underline{0.076} & \textbf{0.042} \\
     & $\nu$ err. & \textbf{0.029} & \textbf{0.020} & \textbf{0.051} \\
     & $\rho$ err. & \textbf{96.79} & \textbf{110.08} & \textbf{87.69} \\
\bottomrule
\end{tabular}
\end{table}

\subsection{Resolution Discussion}
Our current model predicts a $64^3$ voxel field. While this resolution is lower than the grids commonly used for high-quality 3D reconstruction, it is often sufficient for current physical-property prediction and downstream simulation settings. For example, PhysGaussian uses an effective simulation resolution around $50^3$, which is lower than our prediction grid. Since many simulators operate at a similar effective resolution, much finer grids do not necessarily lead to proportional gains in simulation quality. Higher resolution may help with very fine structures, but also incurs significant GPU memory cost; we therefore leave higher-resolution prediction as future work.

\section{\modelname Architecture}
\subsection{Overview}
We propose a text-conditioned 3D U-Net architecture designed to predict continuous physical material parameters (and optionally discrete material segmentation) from dense semantic voxel grids. To effectively fuse high-dimensional visual features with linguistic cues, our architecture is built upon three core components: (1) a \textbf{Feature Projector} for dimensionality reduction of the input semantic features; (2) a hierarchical \textbf{3D U-Net Backbone} that preserves fine-grained spatial topologies and processes geometric structures across multiple resolutions; and (3) a \textbf{FiLM-based Conditioning Module} (Feature-wise Linear Modulation) that dynamically injects text embeddings into the residual blocks, ensuring that the semantic priors guide the spatial feature extraction at every resolution level.

\subsection{Feature Projector}
The feature projector is utilized to map the high-dimensional semantic input features to the network's base conditioning dimension before spatial processing:

\begin{itemize}
    \item \textbf{Input features:} The model processes dense semantic voxel grids, specifically supporting:
    \begin{itemize}
        \item CLIP / OpenCLIP features: $ \mathbf{F} \in \mathbb{R}^{N \times 768 \times D \times H \times W} $ , where $D=H=W=64$
    \end{itemize}
    \item \textbf{Projection:} These 768-dimensional semantic features are projected to a unified base dimension of 32 channels using a feature projector with a hidden dimension of 128. The projector consists of three sequential blocks, each comprising a 3D Convolution, Group Normalization, and a SiLU activation function.
\end{itemize}

\subsection{3D UNet Architecture}
We employ a U-Net architecture ~\cite{shen2023F3RM, le2025pixiefastgeneralizablesupervised} operating on 3D feature grids of shape $ \mathbb{R}^{N \times 32 \times 64 \times 64 \times 64} $. The network follows a standard encoder-decoder structure with skip connections, using a base channel dimension of 64 and channel multipliers of [1, 1, 2, 4] across four resolution levels.

The encoder begins with a 3D convolution that projects the 32-dimensional input features to 64 channels. The encoder then processes features through four resolution levels, each containing three residual blocks. The first two levels maintain 64 channels while progressively reducing spatial dimensions from $ 64^3 $ to $ 32^3 $. The subsequent levels double the channel count at each downsampling step, reaching 128 channels at $ 16^3 $ resolution and 256 channels at $ 8^3 $ resolution. Downsampling between levels is performed using strided 3D convolutions with stride 2.

At the bottleneck, the network processes the lowest resolution features through a sequence of a residual block, an attention block, and another residual block, all operating at $ 8^3 $ spatial resolution with 256 channels. Note that in our implementation, attention blocks are disabled by setting attention resolutions to empty.

The decoder symmetrically reverses the encoder path, utilizing skip connections from corresponding encoder levels. Upsampling is achieved through nearest-neighbor interpolation with a scale factor of 2, followed by 3D convolution. Each decoder level matches the channel dimensions and number of residual blocks of its corresponding encoder level.

Each residual block follows the formulation:
$ \text{ResBlock}(\mathbf{x}) = \mathbf{x} + f(\mathbf{x}) $
where $ f $ consists of layer normalization, LeakyReLU activation with a negative slope of 0.02, 3D convolution with kernel size 3, another layer normalization and activation, dropout, and a final zero-initialized 3D convolution. When input and output channels differ, the skip connection employs a $ 1 \times 1 \times 1 $ convolution for channel matching.

The final output layer applies layer normalization, LeakyReLU activation, and a 3D convolution that projects to either 8 channels for discrete material classification or 3 channels for continuous material parameter regression.

\subsection{FiLM-Conditioned 3D U-Net Architecture}
\label{sec:film_unet}

To inject the text-driven semantic priors into the spatial processing pipeline, we extend the base 3D U-Net into a FiLM-conditioned architecture. This is achieved by embedding Feature-wise Linear Modulation (FiLM) layers within the residual blocks across all resolution levels of the network.

\paragraph{Conditioning Pathway}
The text embedding $ \mathbf{e}_{\text{text}} $ extracted from the frozen OpenCLIP text encoder serves as the global conditioning vector. For each FiLM-integrated layer at the $i$-th resolution level (with channel dimension $ C_i$ $\in$ \allowbreak $\{64,  128, 256\} $), we employ a dedicated two-layer Multi-Layer Perceptron (MLP) with SiLU activation to project $ \mathbf{e}_{\text{text}} $ into the layer-specific modulation parameters: the scaling factor $ \gamma_i \in \mathbb{R}^{C_i} $ and the shifting factor $ \beta_i \in \mathbb{R}^{C_i} $.

\paragraph{FiLM-Integrated Residual Blocks}
We modify the standard residual block to incorporate the text-guided modulation. Recall the original residual formulation $ \text{ResBlock}(\mathbf{x}) = \mathbf{x} + f(\mathbf{x}) $. In our FiLM U-Net, the transformation function $ f $ is augmented to apply the channel-wise affine transformation.

Specifically, the input feature map first passes through layer normalization, a LeakyReLU activation (negative slope 0.02), and a $ 3 \times 3 \times 3 $ 3D convolution. Let this intermediate feature map be $ \mathbf{F}_i $. The FiLM modulation is then applied directly to this feature map:
$ \mathbf{F}_i^{\prime} = \gamma_i(\mathbf{e}_{\text{text}}) \cdot \mathbf{F}_i + \beta_i(\mathbf{e}_{\text{text}}) $
where the multiplication and addition are broadcasted spatially across the 3D voxel grid.

Following the modulation, the conditioned feature map $ \mathbf{F}_i^{\prime} $ is processed by the subsequent layer normalization, LeakyReLU activation, dropout, and the final zero-initialized 3D convolution before being added to the skip connection $ \mathbf{x} $.

\paragraph{Multi-Scale Modulation}
By replacing the standard residual blocks in the encoder, bottleneck, and decoder with these FiLM-integrated blocks, the network achieves multi-scale text-driven modulation. This ensures that the linguistic cues dynamically guide the feature extraction from the highest spatial resolution ($ 64^3 $) down to the deepest semantic bottleneck ($ 8^3 $). Consequently, the network effectively aligns the 3D visual geometry with the specified physical attributes, enabling accurate continuous material parameter regression.

\subsection{Comparison with Other Architectures}

\begin{table}[t]
\centering
\caption{\textbf{Ablation Study of different model architectures.} All models are trained on a subset of \dataset (2,700 train / 1,000 test) for 30 epochs. The results demonstrate that the high-resolution input information is strong enough to generate accurate physics parameters. Architectures that require spatial downsampling or average pooling of the input (\eg, to accommodate sequence lengths) suffer a significant drop in predictive capability.}
\label{tab:architecture_ablation}
\begin{tabular}{lccc}
\toprule
\textbf{Architecture} & \textbf{$ \log E $ err $ \downarrow $} & \textbf{$ \nu $ err $ \downarrow $} & \textbf{$ \log \rho $ err $ \downarrow $} \\
\midrule
MLP (concat input) & 0.316 & 0.245 & 0.298 \\
Transformer ($16^3$) & 0.142 & 0.187 & 0.182 \\
VisualRWKV ($ 16^3 $) & 0.149 & 0.193 & 0.181 \\
VisualRWKV ($ 32^3 $) & 0.106 & 0.124 & 0.125 \\
\textbf{FiLM U-Net (Ours)} & \textbf{0.054} & \textbf{0.061} & \textbf{0.088} \\
\bottomrule
\end{tabular}
\end{table}

To evaluate the efficacy of our FiLM-conditioned 3D U-Net, we compare it against alternative sequence-based architectures, including a 3D Transformer and VisualRWKV, as detailed in Table~\ref{tab:architecture_ablation}. Surprisingly, despite the strong global modeling capabilities typically associated with attention mechanisms, our lightweight FiLM U-Net achieves significantly lower errors across all predicted physical parameters ($ \log E $, $ \nu $, and $ \log \rho $). We attribute this performance gap to the architectural differences in handling 3D spatial topologies and computational efficiency.

\paragraph{Implementation Details and the Cost of Input Pooling.}
To adapt the 3D voxel inputs for sequence-based models, specific architectural compromises are required. For the \textbf{Transformer} baseline, feeding the original $ 64^3 $ voxel grid directly is computationally intractable due to the $ O(N^2) $ complexity of global self-attention (which would yield a sequence length of 262,144 tokens). Therefore, the input must first be spatially downsampled (via average pooling or strided convolutions) to a $ 16^3 $ resolution before being flattened into a 1D sequence of 4,096 tokens. The \textbf{VisualRWKV} baseline, utilizing a linear RNN formulation with $ O(N) $ complexity, can afford a higher spatial resolution of $ 32^3 $ (32,768 tokens), but still requires aggressive input downsampling and 1D flattening. As highlighted in our results, this forced spatial pooling prematurely discards crucial fine-grained geometric information, directly dropping the model's predictive capability.

\paragraph{Preserving 3D Topology vs. Re-learning Geometric Relations.}
A fundamental flaw in applying sequence models to this task lies in how they process spatial structures. The OpenCLIP features utilized in our pipeline inherently capture rich visual priors learned from 2D images. When these features are lifted and aggregated into a 3D voxel grid, they establish a highly structured geometric and topological representation of the object. Flattening this 3D grid into a 1D sequence forces the Transformer and VisualRWKV to completely re-learn these pre-established spatial relationships and 3D topologies from scratch. Given our relatively small training dataset (2,700 samples), demanding the network to rediscover complex geometric priors is exceedingly difficult and highly prone to overfitting. In contrast, the 3D convolutions in our U-Net natively operate on the grid, preserving and directly exploiting this inherent 3D topology.

\paragraph{Resolution Bottleneck and Multi-Scale Processing.}
Because sequence baselines suffer from the aforementioned resolution bottlenecks ($ 16^3 $ or $ 32^3 $), they struggle to predict physics parameters that rely heavily on micro-structures and volumetric thickness. Our FiLM U-Net circumvents this bottleneck by natively processing the high-resolution $ 64^3 $ voxel grids. Through its hierarchical encoder-decoder structure, the network preserves fine-grained geometry at the $ 64^3 $ level while extracting deep semantics at the $ 8^3 $ bottleneck. Combined with multi-scale FiLM modulation, the network successfully injects text conditioning at all geometric levels, leading to superior regression accuracy without destroying the underlying spatial structure.

\section{Simulation Details}

\subsection{\acl{mpm}}
\label{appx:mpm}
\paragraph{Overview}
We use an explicit MPM. Particles (also the 3D Gaussian splats) carry

\begin{equation}
m_p, \; V_p^0, \; \mathbf{X}_p, \; \mathbf{x}_p, \; \mathbf{v}_p, \; \mathbf{F}_p, \; \mathbf{A}_p, \; \mathbf{a}_p, \; \mathbf{\sigma}_p, \; \mathbf{C}_p.
\label{eq:A1}
\end{equation}
In summary, given the 3D GS of a static scene $\{X_p, A_p, \sigma_p, C_p\}$, we use simulation to dynamize the scene by evolving these Gaussians to produce dynamic Gaussians $\{x_p(t), a_p(t), \sigma_p, C_p\}$. Here, \( \mathbf{X}_p \) is the initial  position, while \( \mathbf{x}_p \) is the current  position that evolves over time with velocity \( \mathbf{v}_p \). Furthermore, \( \mathbf{A}_p \) is the static covariance of the initial Gaussian; the dynamic covariance \( \mathbf{a}_p \) is derived at each step; \( \mathbf{F}_p \) is the deformation gradient used to calculate \( \mathbf{a}_p \); and the opacity \( \sigma_p \) and SH coefficient magnitudes \( \mathbf{C}_p \) are considered time-invariant.

\subsubsection{The Material Point Method (MPM) Algorithm Steps}

The Material Point Method (MPM) algorithm iteratively transfers data between particles and a background grid. A single time step can be broken down into the following three main stages.

\subsubsection{Particle-to-Grid Transfer (P2G)}
In the first stage, information is transferred from the Lagrangian particles to the nodes of the Eulerian grid. This process, known as rasterization, effectively creates a grid-based snapshot of the continuum's state. For each particle \(p\), its mass \(m_p\) and momentum \(\mathbf{p}_p = m_p \mathbf{v}_p\) are interpolated and added to the surrounding grid nodes \(i\). This is done using interpolation functions \(N_{ip}\) (also known as shape functions), which depend on the particle's position relative to the grid.

The nodal mass \(m_i\) and nodal momentum \(\mathbf{p}_i\) are computed as follows:

\begin{equation}
    m_i = \sum_p m_p N_{ip}
    \label{eq:A2}
\end{equation}
\begin{equation}
    \mathbf{p}_i = \sum_p m_p \mathbf{v}_p N_{ip}.
    \label{eq:A3}
\end{equation}
From the nodal momentum and mass, the initial nodal velocity is found: \(\mathbf{v}_i = \mathbf{p}_i / m_i\), provided \(m_i > 0\).

\subsubsection{Grid Update}
This stage contains the core physics computations, which are performed entirely on the grid. First, forces acting on each grid node are calculated. These forces are typically composed of two parts:
\begin{itemize}[leftmargin=*]
    \item \textbf{Internal forces} \(\mathbf{f}_i^{\text{internal}}\), which arise from the material's stress. These are computed by transferring particle stress information (derived from the deformation gradient \(\mathbf{F}_p\)) back to the grid.
    \item \textbf{External forces} \(\mathbf{f}_i^{\text{external}}\), such as gravity or user-defined interactions.
\end{itemize}
The total force on a node is \(\mathbf{f}_i = \mathbf{f}_i^{\text{internal}} + \mathbf{f}_i^{\text{external}}\).

With the total force, the grid node velocities are updated over the time step \(\Delta t\) using an explicit time integration scheme (\eg, Forward Euler):
\begin{equation}
    \mathbf{v}_i^{n+1} = \mathbf{v}_i^n + \Delta t \frac{\mathbf{f}_i}{m_i}.
    \label{eq:A4}
\end{equation}
Boundary conditions, such as collisions with obstacles, are also enforced on the grid during this stage by modifying the nodal velocities.

\subsubsection{Grid-to-Particle Transfer (G2P)}
Finally, the updated kinematic information is transferred from the grid back to the particles. This stage, often called the "gather" step, updates the Lagrangian particles' state using the newly computed fields on the Eulerian grid, preparing them for the next time step. This process involves updating each particle's velocity, its deformation gradient, and finally its position.

First, the particle's velocity \(\mathbf{v}_p\) is updated by interpolating the new velocities \(\mathbf{v}_i^{n+1}\) from the surrounding grid nodes. This is essentially a weighted average, using the same interpolation functions \(N_{ip}\) as the P2G step:
\begin{equation}
    \mathbf{v}_p^{n+1} = \sum_i \mathbf{v}_i^{n+1} N_{ip}.
    \label{eq:A5}
\end{equation}
This update can be a pure Particle-In-Cell (PIC) update, or it can be combined with the particle's previous velocity in a FLIP (Fluid-Implicit-Particle) scheme to reduce numerical dissipation.

Simultaneously, the particle's deformation gradient \(\mathbf{F}_p\), which tracks the local rotation and strain of the material, must also be updated. This is done by first computing the velocity gradient \(\nabla \mathbf{v}\) at the particle's position, which is also interpolated from the grid node velocities:
\begin{equation}
    \nabla \mathbf{v}_p = \sum_i \mathbf{v}_i^{n+1} \nabla N_{ip}^T.
\end{equation}
This gradient is then used to advance the deformation gradient forward in time:
\begin{equation}
    \mathbf{F}_p^{n+1} = \left( \mathbf{I} + \Delta t \, \nabla \mathbf{v}_p \right) \mathbf{F}_p^n,
\end{equation}
where \(\mathbf{I}\) is the identity matrix. This update is crucial for correctly computing material stress in the next time step.

Lastly, with the new velocity \(\mathbf{v}_p^{n+1}\) computed, the particle's position \(\mathbf{x}_p\) is updated as:
\begin{equation}
    \mathbf{x}_p^{n+1} = \mathbf{x}_p^n + \Delta t \, \mathbf{v}_p^{n+1}.
\end{equation}
Once all particles have been updated, the information on the background grid is no longer needed and is typically reset or discarded. The simulation is now ready to begin the next time step with the P2G phase.
\begin{figure}[t!]
  \centering
  \includegraphics[width=0.99\textwidth]{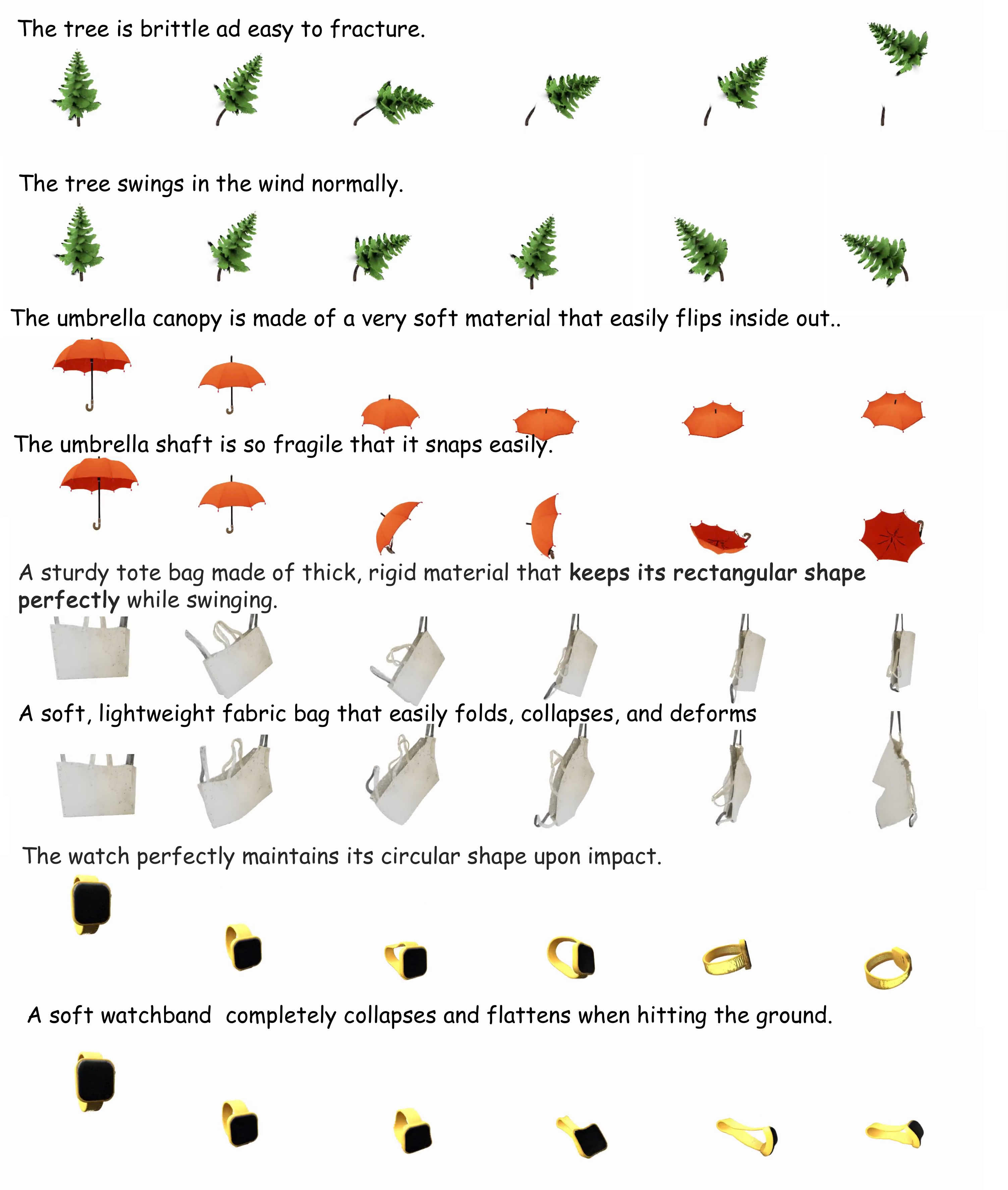}
  \caption{\textbf{More \ac{gs} simulation results.}}
  \label{fig:dataset_example}
\end{figure}

\begin{figure}[t!]
  \centering
  \includegraphics[width=0.99\textwidth]{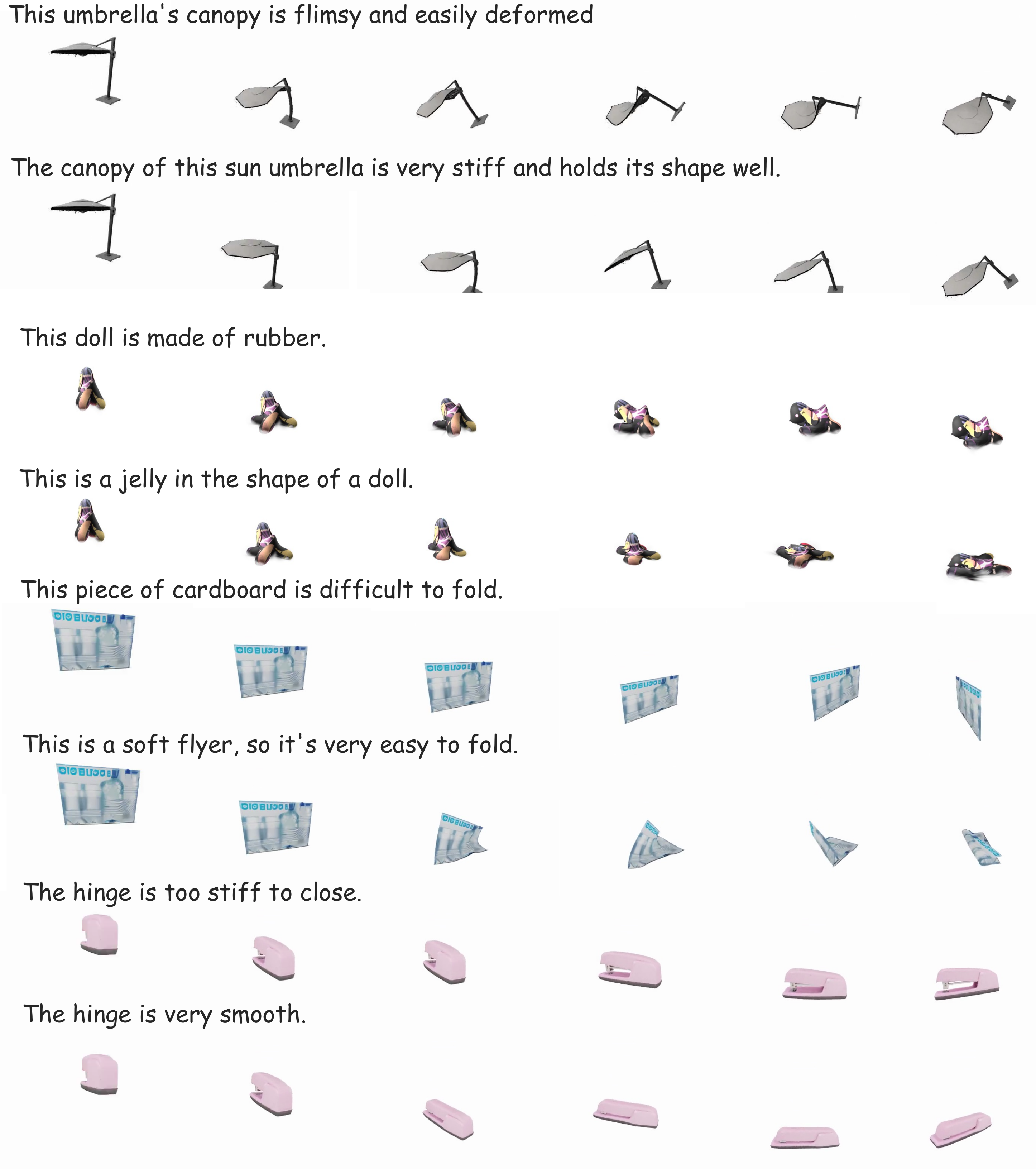}
  \caption{\textbf{More \ac{gs} simulation results.}}
  \label{fig:dataset_example2}
\end{figure}

\begin{figure}[t!]
  \centering
  \includegraphics[width=0.99\textwidth]{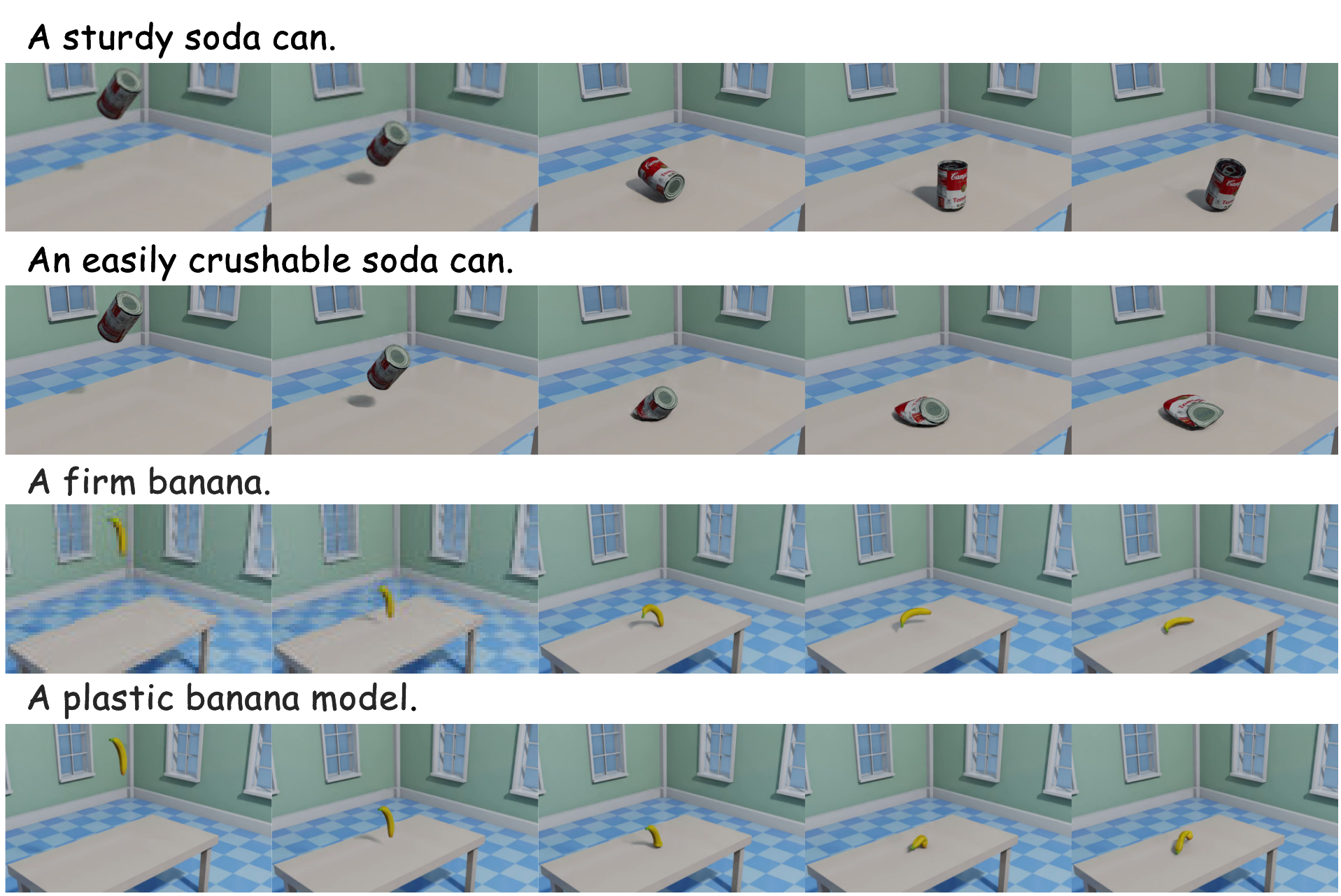}
  \caption{\textbf{More mesh simulation results.}}
  \label{fig:mesh_sim}
\end{figure}
\subsubsection{Evolution of 3D Gaussian Properties via Continuum Mechanics}

This approach outlines a method for animating 3D GS by treating them as discrete particles within a physics-based system governed by continuum mechanics. The primary goal is to evolve a static scene, defined by initial properties, into a dynamic state for rendering.

The evolution of the key Gaussian properties for each time step is as follows:

\begin{itemize}[leftmargin=*]
    \item \textbf{Position Evolution (Mean):} The Gaussian's center, or mean, is its world-space position \(\mathbf{x}_p\). This is updated using the particle's velocity \(\mathbf{v}_p\), which is determined by the physical simulation, via explicit time integration:
    \begin{equation}
        \mathbf{x}_p^{n+1} = \mathbf{x}_p^n + \Delta t \, \mathbf{v}_p.
    \end{equation}

    \item \textbf{Shape Evolution (Covariance):} The dynamic world-space covariance \(\mathbf{a}_p\), which defines the Gaussian's shape and size, is computed directly from the deformation gradient \(\mathbf{F}_p\). The deformation gradient describes the local deformation of the material around the particle. It maps the initial, undeformed shape (defined by the material-space covariance \(\mathbf{A}_p\)) to its current, deformed configuration:
    \begin{equation}
        \mathbf{a}_p(t) = \mathbf{F}_p(t) \mathbf{A}_p \mathbf{F}_p(t)^T.
    \end{equation}

    \item \textbf{Orientation Evolution (for Rendering):} To correctly render anisotropic appearances (\eg, using Spherical Harmonics), the particle's orientation must be tracked. The rotation component \(\mathbf{R}_p\) is extracted from the deformation gradient, typically via polar decomposition (\(\mathbf{F}_p = \mathbf{R}_p \mathbf{S}_p\)). This rotation is then applied to the appearance model during rendering.

    \item \textbf{Time-Invariant Properties:} Visual attributes such as opacity \(\sigma_p\) and material-space appearance coefficients (\eg, Spherical Harmonics, \(\mathbf{C}_p\)) are considered intrinsic material properties. They are typically held constant throughout the simulation.
\end{itemize}

\subsection{\acl{fem}}
\label{appx:fem}
\paragraph{Overview}
Unlike the hybrid particle-grid approach of MPM, the \acl{fem} operates directly on a connected mesh structure, typically composed of tetrahedral or hexahedral elements. The continuum is discretized into vertices (nodes) and elements. The vertices carry the primary kinematic state:
$ m_i, \; \mathbf{X}_i, \; \mathbf{x}_i, \; \mathbf{v}_i, \; \mathbf{f}_i $
where $ m_i $ is the nodal mass, $ \mathbf{X}_i $ and $ \mathbf{x}_i $ are the initial and current positions, $ \mathbf{v}_i $ is the velocity, and $ \mathbf{f}_i $ is the force. The elements $ e $ carry the deformation state, notably the deformation gradient $ \mathbf{F}_e $. FEM is widely adopted for simulating deformable solids due to its accuracy in preserving volume and capturing rich elastic behaviors without the numerical dissipation often seen in grid-based transfers.

\subsubsection{The Finite Element Method  Algorithm Steps}
An explicit FEM time step computes deformations and forces directly on the mesh topology. The process generally involves the following stages:

\paragraph{Deformation Gradient Computation}
For each tetrahedral element $ e $ consisting of four vertices, the deformation gradient $ \mathbf{F}_e $ is computed by comparing the current shape matrix $ \mathbf{D}_s $ to the reference shape matrix $ \mathbf{D}_m $:
$ \mathbf{F}_e = \mathbf{D}_s \mathbf{D}_m^{-1} $
Here, $ \mathbf{D}_s $ is formed by the edge vectors of the tetrahedron in the current configuration (\eg, $ \mathbf{x}_1 - \mathbf{x}_0 $, $ \mathbf{x}_2 - \mathbf{x}_0 $, $ \mathbf{x}_3 - \mathbf{x}_0 $), and $ \mathbf{D}_m^{-1} $ is the precomputed inverse of the edge vectors in the rest configuration.

\paragraph{Force Computation}
Based on the deformation gradient $ \mathbf{F}_e $, the internal stress (\eg, the first Piola-Kirchhoff stress $ \mathbf{P}_e $) is evaluated using a chosen hyperelastic constitutive model, such as Neo-Hookean or Corotated Linear Elasticity. The internal elastic forces are then distributed to the vertices of the element:
$ \mathbf{f}_i^{\text{internal}} = \sum_{e \in \Omega_i} \text{Vol}_e \mathbf{P}_e \frac{\partial \mathbf{F}_e}{\partial \mathbf{x}_i} $
where $ \text{Vol}_e $ is the rest volume of the element, and $ \Omega_i $ denotes the set of elements adjacent to vertex $ i $. Total nodal forces are accumulated by adding external forces (\eg, gravity or collisions):
$$ \mathbf{f}_i = \mathbf{f}_i^{\text{internal}} + \mathbf{f}_i^{\text{external}} $$

\paragraph{Time Integration}
With the total nodal forces computed, the vertex velocities and positions are updated using a time integration scheme (\eg, Symplectic Euler):
$$ \mathbf{v}_i^{n+1} = \mathbf{v}_i^n + \Delta t \frac{\mathbf{f}_i}{m_i} $$
$$ \mathbf{x}_i^{n+1} = \mathbf{x}_i^n + \Delta t \, \mathbf{v}_i^{n+1} $$

\subsubsection{FEM in Modern GPU-Accelerated Simulators}
Recent advancements in physics simulation have heavily leveraged GPU acceleration to simulate high-fidelity FEM meshes at interactive or faster-than-real-time rates. Platforms such as NVIDIA Omniverse and Isaac Gym provide robust, hardware-optimized implementations of FEM for soft body dynamics.

In these environments, the FEM pipeline is parallelized across thousands of CUDA cores. Isaac Gym, in particular, allows for massive parallelization of environments (tensorized simulation), which is highly beneficial for training Reinforcement Learning (RL) agents. By representing deformable objects as tetrahedral FEM meshes within these platforms, researchers can achieve physically accurate sim-to-real transfer for robotic manipulation tasks involving soft materials. The underlying solvers in these engines often employ advanced techniques, such as XPBD (Extended Position-Based Dynamics) or GPU-based Newton-descent solvers, to maintain stability even under large deformations and complex collision scenarios.

\end{document}